\documentclass[a4paper]{cas-dc}
\usepackage[numbers,sort&compress]{natbib}
\usepackage{multirow,colortbl,tabularx,booktabs,xcolor}
\usepackage{amsmath,amssymb,mathtools,bm}
\usepackage{algorithm,algpseudocode}
\usepackage{graphicx,adjustbox,microtype}
\usepackage{placeins}
\newcommand{\AlignedFloatBarrier}{\begingroup\renewcommand{\newpage}{\pagebreak[4]}\FloatBarrier\endgroup}
\usepackage{stfloats}
\usepackage{flushend}
\makeatletter
\ExplSyntaxOn
\exp_args:Nc \patchcmd {environment~Abstract~code}
  {\hbox_to_wd:nn {\z@} {\box \g_stm_key_box}}
  {\hbox_to_wd:nn {\z@} {\box \g_stm_key_box \hss}}
  {} {\PackageError{flowclip}{Cannot patch CAS keyword overlay}{}}
\cs_new_eq:NN \flowclip_original_dbflt:n \@dbflt
\cs_set_protected:Npn \@dbflt #1
  {
    \str_if_eq:nnTF {#1}{table}
      {
        \@ifnextchar [
          { \@xdblfloat {#1} }
          { \use:x { \exp_not:N \@xdblfloat {#1} [ \l_tbl_pos_tl ] } }
      }
      {
        \str_if_eq:nnTF {#1}{figure}
          {
            \@ifnextchar [
              { \@xdblfloat {#1} }
              { \use:x { \exp_not:N \@xdblfloat {#1} [ \l_fig_pos_tl ] } }
          }
          { \flowclip_original_dbflt:n {#1} }
      }
  }
\ExplSyntaxOff
\makeatother
\usepackage[nameinlink]{cleveref}
\definecolor{prrefblue}{RGB}{0,70,160}
\hypersetup{colorlinks=true,linkcolor=prrefblue,citecolor=prrefblue,urlcolor=prrefblue,hypertexnames=false}
\graphicspath{{./}}
\crefname{figure}{Fig.}{Figs.}
\Crefname{figure}{Fig.}{Figs.}
\crefname{table}{Tab.}{Tabs.}
\Crefname{table}{Tab.}{Tabs.}
\crefname{algorithm}{Algorithm}{Algorithms}
\Crefname{algorithm}{Algorithm}{Algorithms}
\crefname{equation}{Eq.}{Eqs.}
\Crefname{equation}{Eq.}{Eqs.}
\crefname{section}{Sec.}{Secs.}
\Crefname{section}{Sec.}{Secs.}
\crefname{subsection}{Sec.}{Secs.}
\Crefname{subsection}{Sec.}{Secs.}
\crefname{subsubsection}{Sec.}{Secs.}
\Crefname{subsubsection}{Sec.}{Secs.}
\newcommand{\vimg}{\bm{v}_I}
\newcommand{\tcap}{\bm{t}_T}
\newcommand{\score}{s}

\newcommand{\cmark}{\ensuremath{\checkmark}}
\newcommand{\xmark}{\ensuremath{\times}}
\DeclareRobustCommand{\frozenicon}{\raisebox{-0.08em}{\includegraphics[height=0.82em]{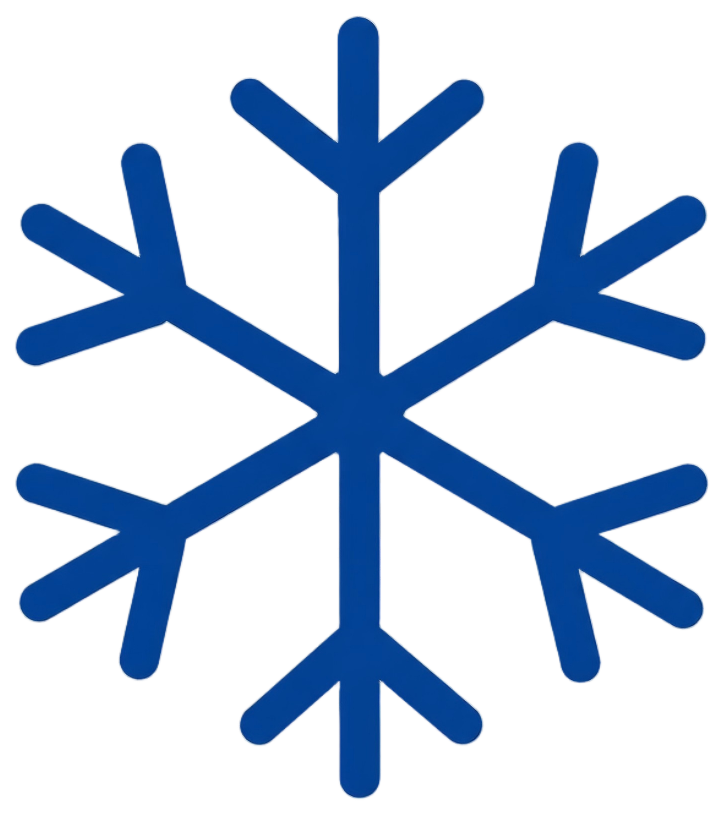}}}
\DeclareRobustCommand{\learnableicon}{\raisebox{-0.08em}{\includegraphics[height=0.82em]{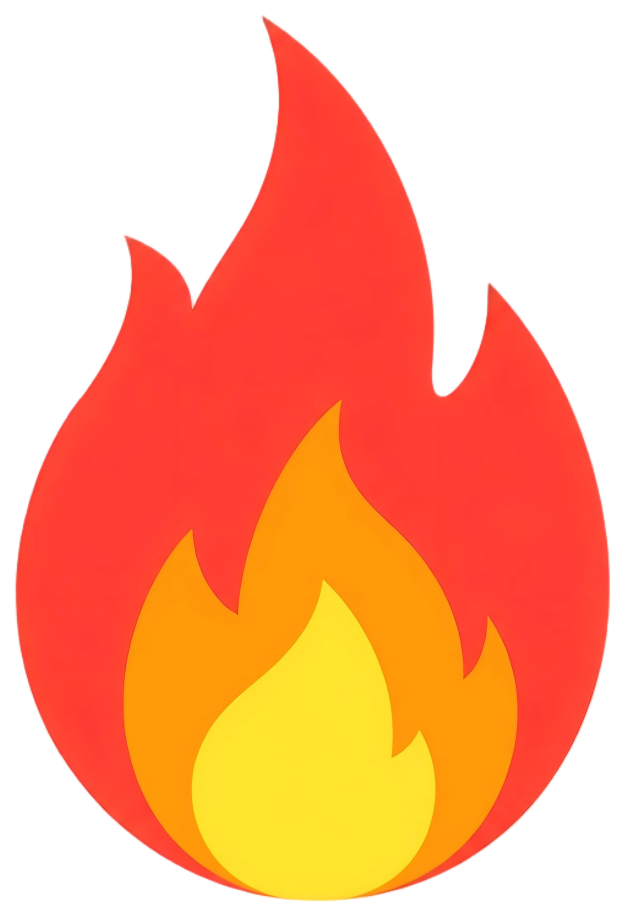}}}
\newcommand{\crepecomicfont}{\sffamily}

\definecolor{trainposgreen}{RGB}{0,125,20}
\definecolor{trainnegorange}{RGB}{220,90,0}

\ExplSyntaxOn
\cs_set_protected:Npn \__make_tbl_caption:nn #1#2
  {
    \l_tbl_align_tl
    \skip_vertical:N \l_tbl_abovecap_skip
    \parbox{\l_tbl_width_dim}
      {\normalfont\rmfamily\footnotesize\rightskip=0pt
       \textbf{#1}\par #2\par\vskip4pt}
    \skip_vertical:N \l_tbl_belowcap_skip
  }
\cs_set_protected:Npn \__make_fig_caption:nn #1#2
  {
    \l_fig_align_tl
    \skip_vertical:N \l_fig_abovecap_skip
    \parbox{\l_fig_width_dim}
      {\normalfont\rmfamily\footnotesize\rightskip=0pt
       \textbf{#1.}\enspace #2\par}
    \skip_vertical:N \l_fig_belowcap_skip
  }
\ExplSyntaxOff


\makeatletter
\AtBeginDocument{\@fleqnfalse\@mathmargin\@centering}
\makeatother

\stmausetup{size=\normalsize,shape=\upshape,weight=\mdseries,color=black,before=3pt,after=6pt}
\ExplSyntaxOn
\RenewDocumentCommand \firstname { }
  {\textcolor{black}{\seq_use:Nn \l_stm_au_seq {~}}}
\RenewDocumentCommand \processAffNum { m }
  {
    \seq_clear:N \l_affmk_seq
    \clist_map_inline:nn {#1}
      {\seq_put_right:Nn \l_affmk_seq {\int_to_arabic:n {##1}}}
  }
\NewCommandCopy \flowclipOriginalStmAddress \stmaddress
\RenewDocumentCommand \stmaddress { }
  {
    \group_begin:
    \cs_set_eq:NN \int_to_alph:n \int_to_arabic:n
    \flowclipOriginalStmAddress
    \group_end:
  }
\ExplSyntaxOff

\ExplSyntaxOn
\cs_set:Npn \__first_head: { }
\cs_set:Npn \__cas_head: { }
\cs_set:Npn \__first_footerline: { }
\cs_set:Npn \__first_foot:
  {\hfil\normalfont\rmfamily\small\thepage\hfil}
\cs_set_eq:NN \__cas_foot: \__first_foot:
\ExplSyntaxOff
\AtBeginDocument{\pagestyle{cas}}

\begin{document}
\shorttitle{Flow-CLIP}
\shortauthors{M. Wei et~al.}
\title[mode=title]{Flow-CLIP: Image-Conditioned Residual Scoring for Compositional Image--Text Matching}
\author[1]{Minghong Wei}
\ead{wmh0517@bupt.edu.cn}
\author[1]{Enguang Zhu}
\ead{2025141266@bupt.cn}
\author[1]{Pu Cao}
\ead{caopu@bupt.edu.cn}
\author[1]{Lu Yang}
\cormark[1]
\ead{soeaver@bupt.edu.cn}
\author[1]{Qing Song}
\ead{priv@bupt.edu.cn}
\affiliation[1]{organization={School of Intelligent Engineering and Automation, Beijing University of Posts and Telecommunications}, city={Beijing}, postcode={100082}, country={China}}
\cortext[1]{Corresponding author.}

\begin{abstract}
Vision--language pre-training has enabled substantial progress in image--text retrieval and zero-shot recognition by learning transferable cross-modal representations. However, sentence-level alignment remains insufficient for compositional matching, where captions can describe the same objects and attributes while differing in how these elements are bound. To address this limitation, we propose Flow-CLIP, which refines global image--text alignment through image-conditioned changes in token--patch evidence. Global Compatibility Calibration (GCC) learns a sentence-level similarity metric, while Image-Conditioned Evidence Refinement (ICER) incorporates visual patch context into token queries and uses the resulting change in local matching evidence to adjust the global score. We further introduce Cross-Image Margin Ranking (CIMR), which compares the same positive--negative caption pair across matched and mismatched images, encouraging a larger ranking margin when the image supports the correct description. Experiments on compositional matching and attribute--object binding benchmarks show that Flow-CLIP improves binding accuracy over pretrained CLIP and learned global alignment through local evidence refinement, while cross-image ranking supervision mitigates reliance on language priors. Code will be released at \href{https://github.com/weiminghong/Flow-CLIP}{\textcolor{red}{link}}.
\end{abstract}

\maketitle
\flushbottom

\begin{figure}[pos=tbp]
	\centering
	\includegraphics[width=\columnwidth]{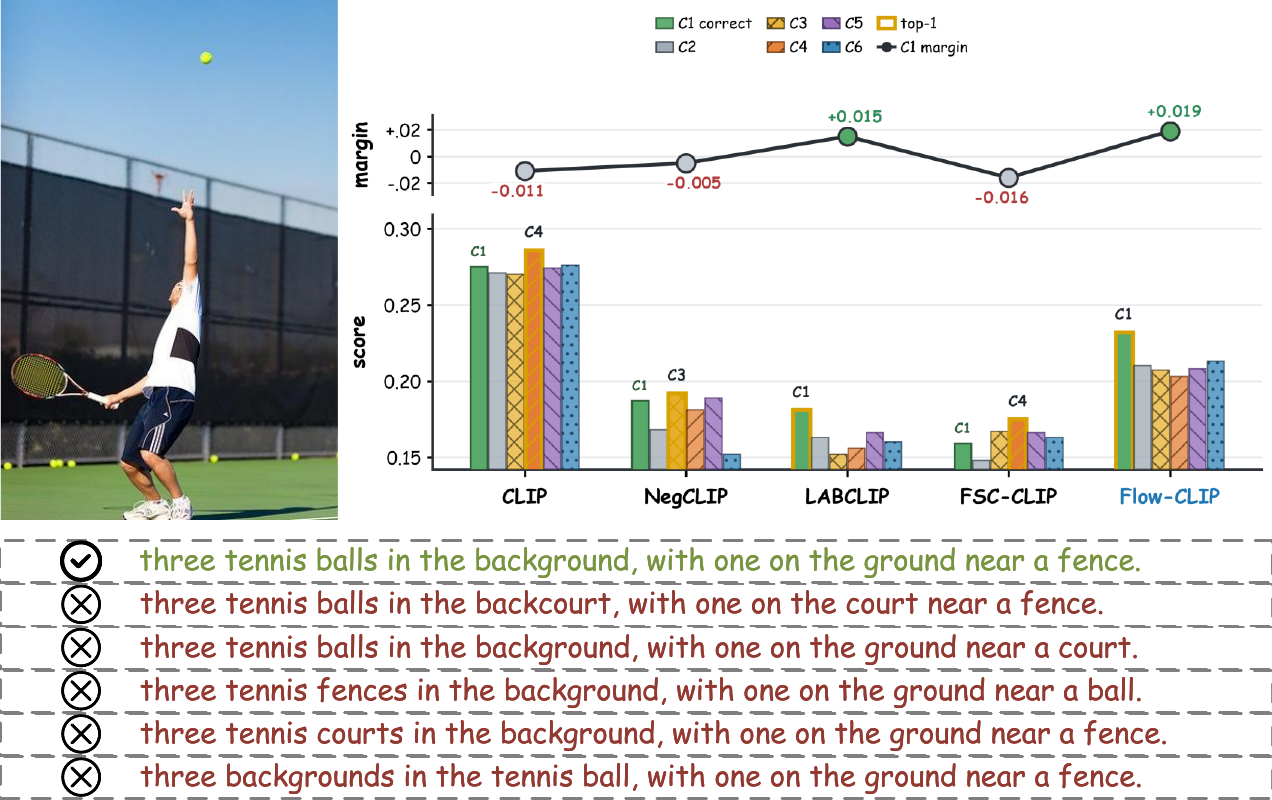}
	\caption{Compositional image--text matching with closely related caption candidates. Bars show compatibility scores, yellow outlines identify the top-ranked caption, and the curve shows the gold-caption margin.}
	\label{fig:global-local-motivation}
\end{figure}

\section{Introduction}\label{sec:introduction}

Vision--language pretraining enables zero-shot recognition and image--text retrieval by learning transferable representations from paired images and descriptions~\citep{radford2021learning,jia2021scaling,li2022blip,zhai2022lit,zhai2023siglip}. Compositional matching requires these representations to distinguish descriptions that name the same objects and attributes but assign or relate them differently. Benchmarks of attribute binding, relations and word order document persistent errors in making these distinctions~\citep{thrush2022winoground,yuksekgonul2023and,zhao2022vlchecklist,parcalabescu2022valse,lewis2024bind,kamath2023whatsup}.

The caption rankings in \Cref{fig:global-local-motivation} illustrate how CLIP-based models~\citep{radford2021learning,yuksekgonul2023and,oh2024preserving} can prefer a hard-negative caption to the positive caption. The alternatives share many words but differ in object identities, attribute assignments or spatial relations. Targeted caption negatives expose these distinctions during training~\citep{yuksekgonul2023and,doveh2023teaching}, while dense descriptions and image--text--graph comparisons add supervision for scene details~\citep{doveh2023dense,singh2023coarse}. FSC-CLIP~\citep{oh2024preserving} combines local hard-negative training with calibrated regularization to improve compositional matching while maintaining general-task performance.

LABCLIP~\citep{koishigarina2025clip} shows that a learned linear transformation of text embeddings can recover attribute--object binding information in cross-modal matching. We use learned global alignment as a baseline to measure the additional contribution of image-conditioned local refinement to attribute--object binding.

Local matching provides the representations needed for such refinement. FILIP~\citep{yao2021filip} and DCSM~\citep{kang2025clip} use token--patch similarities, ComCLIP~\citep{jiang2024comclip} and OC-CLIP~\citep{assouel2024oc} introduce structured entity scores, and FLAIR~\citep{xiao2025flair} constructs caption-conditioned visual representations. These methods aggregate local similarities, compare structured entities or condition visual features on text. We use local interactions to measure how image context changes support for a caption, providing a correction to its calibrated global score.

\begin{figure*}[pos=tbp]
	\centering
	\includegraphics[width=\textwidth]{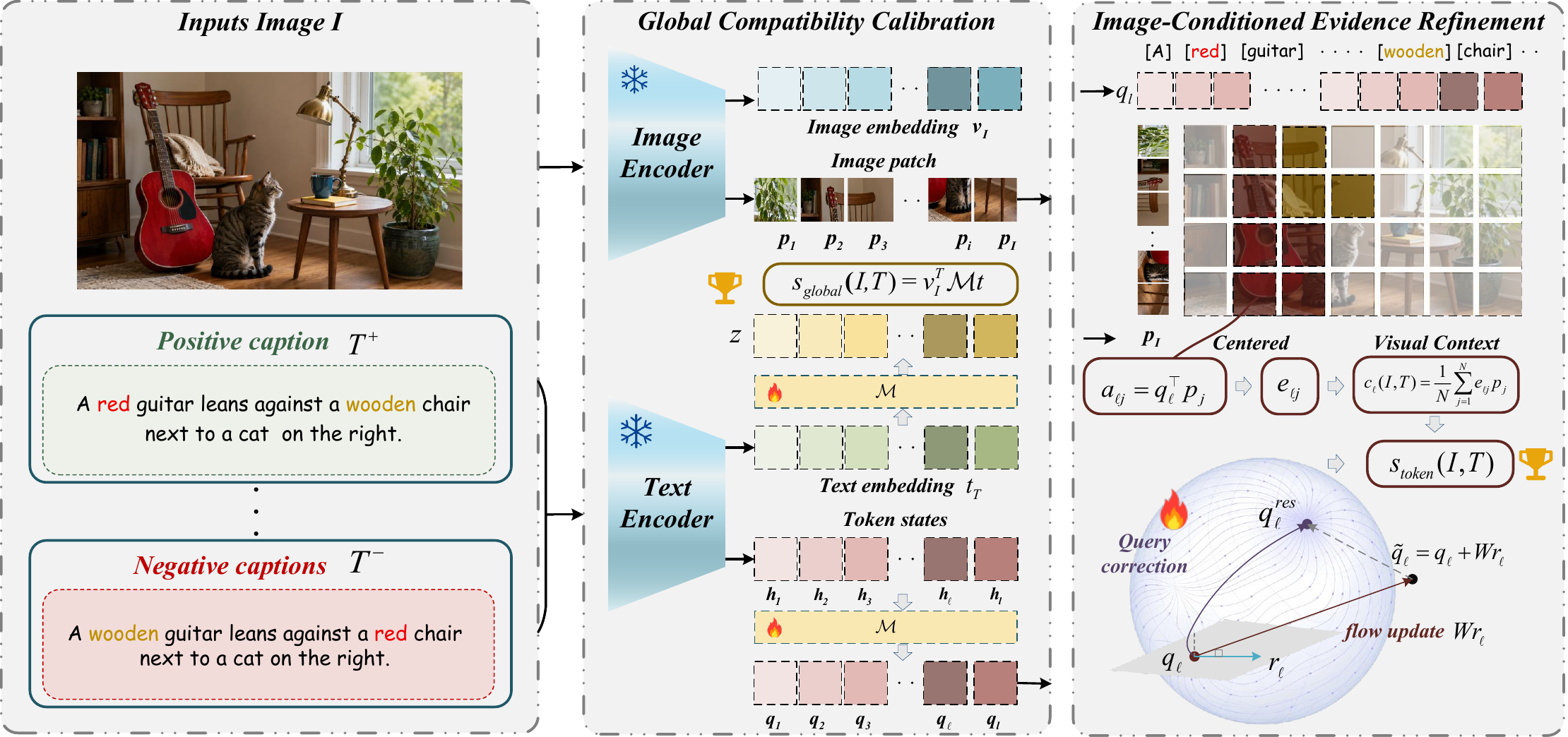}
	\caption{Flow-CLIP combines Global Compatibility Calibration (GCC) with Image-Conditioned Evidence Refinement (ICER). The global and token maps produce normalized representations. Centered patch support forms visual context, whose projection supplies the input to the learned query correction. The normalized corrected query and its original query yield a signed difference in pooled evidence, which is averaged over valid tokens and added to global compatibility. \protect\frozenicon\ and \protect\learnableicon\ denote frozen and learned components.}
	\label{fig:method-overview}
\end{figure*}

An incorrect attribute assignment can receive local support when its individual objects and attributes are present in the image. Strong token--patch matches alone therefore leave open whether the proposed assignment is supported. We use the image's patch context to refine the token queries that assess this assignment, then compare the refined and original queries on the same patches. The original query supplies a reference for measuring the correction's effect, yielding a signed adjustment that can strengthen or weaken the global compatibility score. To model this image-conditioned change, we propose Flow-CLIP, which combines \textbf{Global Compatibility Calibration (GCC)} with \textbf{Image-Conditioned Evidence Refinement (ICER)}, as illustrated in \Cref{fig:method-overview}. GCC learns sentence-level alignment, and ICER supplies the local adjustment through context-dependent query correction. Both components are jointly trained for caption ranking with frozen encoders.

Full-caption hard negatives supervise caption ranking on matched images, following NegCLIP~\citep{yuksekgonul2023and}. The construction of these alternatives also matters: SugarCrepe~\citep{hsieh2023sugarcrepe} identifies plausibility and fluency biases that allow models to distinguish positive descriptions from negative alternatives without the image. Language priors can consequently favor plausible descriptions even when their attribute assignments are unsupported by the visual input. Supervision on matched pairs does not explicitly constrain how this preference should change across images. A caption preference that persists after replacing the image can satisfy the matched-pair objective without reflecting the visual evidence for the attribute assignment. We introduce \textbf{Cross-Image Margin Ranking (CIMR)} to compare the same positive--negative caption margin on the matched image and one donor shared by both captions. By encouraging a larger margin on the matched image, this objective rewards a change in preference tied to visual input while holding the captions and their linguistic cues fixed. GCC and ICER are optimized jointly with the hard-negative and CIMR objectives.

Independently trained global-only heads account for much of the caption-order improvement, while training with the local branch yields additional attribute--object binding gains. Paired predictions identify corrected binding errors and new spatial errors. Image-replacement tests measure changes in caption preference when the visual match is removed. Comparisons across training sources and encoders show how binding gains vary alongside changes in spatial and retrieval accuracy.

The contributions of this work are:
\begin{itemize}
    \item We propose Flow-CLIP, combining Global Compatibility Calibration (GCC) and Image-Conditioned Evidence Refinement (ICER) with frozen encoders. The local branch refines token queries using visual patch context and adjusts global compatibility by the signed change in pooled patch evidence.
    \item We introduce Cross-Image Margin Ranking (CIMR), which compares the same positive--negative caption margin on matched and shared donor images. This objective encourages caption preference to depend on the matched image using full-caption supervision.
    \item Our experiments separate the binding gains of local refinement from the sentence-level gains of global calibration and identify the contribution of cross-image supervision to mitigating reliance on language priors.
\end{itemize}

\section{Related Work}\label{sec:related-work}

\subsection{Compositional Evaluation}

Compositional benchmarks evaluate image--text matching through controlled semantic changes. Winoground~\citep{thrush2022winoground} pairs two images with captions that share their words but differ in composition, requiring scores to recover both image--caption assignments. ARO~\citep{yuksekgonul2023and} separates attribution, relation and caption-order judgments. VL-CheckList~\citep{zhao2022vlchecklist} organizes object, attribute and relation edits, while VALSE~\citep{parcalabescu2022valse} evaluates linguistic phenomena using positive descriptions and edited foils. What's Up?~\citep{kamath2023whatsup} focuses on spatial relations. These distinctions matter for local scoring: recognizing the concepts present in an image can support object discrimination, whereas attribute reassignment and relation reversal require resolving how those concepts are associated.

CLEVR~\citep{johnson2017clevr} and PUG~\citep{bordes2023pug} provide rendered scenes with known attribute--object assignments. Swapping those assignments produces caption alternatives that preserve the scene's object and attribute inventory, isolating the assignment distinction relevant to token-level refinement. CREPE~\citep{ma2023crepe} tests systematicity and productivity with candidate sets of increasing compositional complexity, and BiVLC~\citep{miranda2024bivlc} evaluates both retrieval directions. These evaluations extend the comparison to competition among multiple descriptions and to the consistency of image--text matching across retrieval directions.

Caption construction determines which cues can support a correct ranking. SugarCrepe~\citep{hsieh2023sugarcrepe} uses fluent semantic edits to reduce linguistic artifacts, and SugarCrepe++~\citep{dumpala2024sugarcrepe++} distinguishes semantic sensitivity from lexical variation. NegBench~\citep{alhamoud2025negation} evaluates negation, whereas Half-Truths~\citep{kargi2026half} adds plausible but incorrect content to an otherwise valid description. Controlling the caption alternatives limits cues introduced by their construction. Image-replacement controls address the complementary question of whether the resulting preference depends on the visual match while keeping those alternatives fixed.

\subsection{Image--Text Matching}

Global scoring compares pooled image and text embeddings, as used in CLIP~\citep{radford2021learning} and ALIGN~\citep{jia2021scaling}. LiT~\citep{zhai2022lit} studies alignment with a locked image encoder. LABCLIP~\citep{koishigarina2025clip} evaluates binding within each modality using linear probes, increasing object counts and conjunctive search, then learns a text-embedding transformation for cross-modal matching. These experiments separate the information represented by each encoder from the information accessible through the cross-modal scoring function.

Work on concept-to-object mappings~\citep{uselis2026binding} further examines how embeddings represent binding functions. The distinction relevant to our comparison is the granularity of the learned alignment: LABCLIP transforms the pooled text embedding, while ICER updates individual token queries using image patches. GCC retains the sentence-level transformation, allowing the experiments to assess the additional effect of training with the local branch under matched conditions.

Fine-grained alignment preserves visual and textual units within the matching score. RegionCLIP~\citep{zhong2021regionclip} learns region--text representations, and FILIP~\citep{yao2021filip} aggregates token-wise maximum similarities for cross-modal late interaction. FLAIR~\citep{xiao2025flair} uses text-conditioned attention pooling over image tokens for detailed retrieval, while Beta-CLIP~\citep{zohra2025beta} develops text-conditioned alignment at multiple granularities. The conditioning differs across these methods: late interaction selects matching responses, whereas text-conditioned pooling changes the visual representation.

DCSM~\citep{kang2025clip} learns a scoring function over dense token--patch cosine similarities, including functional rows for spatial relationship words. ICER scores the change in pooled support after query correction.

Structured matching specifies the semantic units used for comparison. ComCLIP~\citep{jiang2024comclip} computes separate scores for subject, object and action subimages, then combines these scores at inference. OC-CLIP~\citep{assouel2024oc} associates scene-graph nodes with visual slots and scores relational compatibility, while ABE-CLIP~\citep{zhang2025abe} introduces attribute-binding operations during inference. VQAScore~\citep{li2024vqascore} scores image--text compatibility through visual question answering. Flow-CLIP uses contextual token and patch states as its local comparison units, without requiring explicit scene-graph or subimage decomposition. Full-caption supervision trains the contribution of these interactions to the final compatibility score.

\subsection{Training Objectives}

Compositional training exposes distinctions that ordinary image--caption pairs may leave implicit. NegCLIP~\citep{yuksekgonul2023and} uses caption perturbations and hard image negatives, and TSVLC~\citep{doveh2023teaching} manipulates language structure to teach attributes, relations and states from paired data. DAC~\citep{doveh2023dense} supplies dense, image-aligned descriptions. Coarse-to-fine image--text--graph learning~\citep{singh2023coarse} decomposes and augments parsed scene graphs to align descriptions of different structural complexities with an image. TripletCLIP~\citep{patel2024tripletclip} and SPARCL~\citep{li2025enhancing} add synthetic visual and textual comparisons. Flow-CLIP trains its global and local terms through the same image--caption score using full-caption hard negatives.

The objective determines which relationships those training examples constrain. CE-CLIP~\citep{zhang2024contrasting} introduces intra-modal contrast and cross-modal hard-negative ranking. FSC-CLIP~\citep{oh2024preserving} combines local hard-negative loss with selective calibrated regularization to balance compositional supervision and general multimodal capabilities. Its design addresses the degradation of pretrained multimodal representations caused by hard-negative fine-tuning. READ-CLIP~\citep{kwon2025readclip} couples token reconstruction with paraphrase alignment, while concept-centric learning~\citep{pham2026no} aligns caption concepts with visual content through auxiliary contrasts. These objectives constrain feature alignment or its preservation during adaptation.

CLIC~\citep{peleg2026advancing} constructs training comparisons from combinations of multiple images and their captions. CIMR evaluates the same caption pair on both the matched image and a shared donor image. Caption identity remains fixed when the two margins are compared, so an additive text-only term cancels and the objective constrains the score change induced by replacing the image.

\section{Method}\label{sec:method}

For an image $I$ and caption $T$, Flow-CLIP adds an ICER local term to GCC global compatibility. The local term measures the signed change in pooled patch support after image-conditioned token-query correction:
\begin{equation}
    \score(I,T)=s_{\mathrm{global}}(I,T)+\alpha s_{\mathrm{token}}(I,T).
    \label{eq:flowclip-score}
\end{equation}
Here $s_{\mathrm{global}}$ denotes global compatibility, $s_{\mathrm{token}}$ denotes the local evidence change, and $\alpha$ scales the local contribution. The main configuration freezes both encoders; encoder-adapted configurations use the same score. Global-$M$ denotes the global-only GCC scorer.

\begin{algorithm}[tbp]
	\caption{Flow-CLIP scoring with GCC and ICER}
	\label{alg:flowclip}
	\small
	\algrenewcommand\algorithmicrequire{\textbf{Input:}}
	\algrenewcommand\algorithmicensure{\textbf{Output:}}
	\algrenewcommand\algorithmicindent{1em}
	\begin{algorithmic}[1]
		\Require Image $I$, caption $T$; image and text encoders; learned mappings $M,M_{\mathrm{tok}},W$; pooling temperature $\tau_p>0$ and local scale $\alpha\geq0$.
		\Ensure Compatibility score $\score(I,T)$.
		\Statex \textit{Feature extraction and GCC}
		\State Extract global image embedding $\vimg$ and patch embeddings $P$ from $I$.
		\State Extract sentence embedding $\tcap$ and projected token states $H$ from $T$.
		\State $\vimg\gets\operatorname{norm}(\vimg)$; $\tcap\gets\operatorname{norm}(\tcap)$.
		\State $\bm{p}_j\gets\operatorname{norm}(\bm{p}_j)$, $j=1,\ldots,N$.
		\State $\bm{h}_\ell\gets\operatorname{norm}(\bm{h}_\ell)$, $\ell=1,\ldots,L$.
		\State Obtain $\mathcal{V}_T$ by excluding special and padding tokens.
		\State $\bm{z}\gets\operatorname{norm}(M\tcap)$.
		\State $s_{\mathrm{global}}\gets\vimg^\top\bm{z}$ using \Cref{eq:global-score}.
		\If{$\mathcal{V}_T=\varnothing$}
		\State \Return $s_{\mathrm{global}}$.
		\EndIf
		\Statex \textit{ICER: image-conditioned token correction}
		\For{$\ell\in\mathcal{V}_T$}
		\State $\bm{q}_\ell\gets\operatorname{norm}(M_{\mathrm{tok}}\bm{h}_\ell)$.
		\State $a_{\ell j}\gets\bm{q}_\ell^\top\bm{p}_j$, $j=1,\ldots,N$.
		\State $\bar a_\ell\gets\frac{1}{N}\sum_{j=1}^{N}a_{\ell j}$.
		\State $e_{\ell j}\gets\max(a_{\ell j}-\bar a_\ell,0)$.
		\State $\bm{c}_\ell\gets\frac{1}{N}\sum_{j=1}^{N}e_{\ell j}\bm{p}_j$.
		\State $\bm{r}_\ell\gets\bm{c}_\ell-(\bm{q}_\ell^\top\bm{c}_\ell)\bm{q}_\ell$.
		\State $\tilde{\bm{q}}_\ell\gets\bm{q}_\ell+W\bm{r}_\ell$.
		\State $\bm{q}^{\mathrm{res}}_\ell\gets\operatorname{norm}(\tilde{\bm{q}}_\ell)$.
		\State $\rho_\ell\gets\mathcal{P}_{\tau_p}(\bm{q}_\ell,P)$ using \Cref{eq:patch-evidence}.
		\State $\rho^{\mathrm{res}}_\ell\gets\mathcal{P}_{\tau_p}(\bm{q}^{\mathrm{res}}_\ell,P)$.
		\State $\Delta_\ell\gets\rho^{\mathrm{res}}_\ell-\rho_\ell$ using \Cref{eq:residual-increment}.
		\EndFor
		\Statex \textit{Score aggregation}
		\State $\displaystyle s_{\mathrm{token}}\gets\frac{1}{|\mathcal{V}_T|}\sum_{\ell\in\mathcal{V}_T}\Delta_\ell$.
		\State \Return $s_{\mathrm{global}}+\alpha s_{\mathrm{token}}$.
	\end{algorithmic}
\end{algorithm}

\subsection{Global Compatibility Calibration}

For an image--caption pair $(I,T)$, the image encoder produces a global embedding $\vimg\in\mathbb{R}^{d}$ and $N$ patch embeddings $P=[\bm{p}_1,\ldots,\bm{p}_N]^\top\in\mathbb{R}^{N\times d}$. The text encoder produces a sentence embedding $\tcap\in\mathbb{R}^{d}$ and projected states for a sequence of $L$ tokens, $H=[\bm{h}_1,\ldots,\bm{h}_L]^\top\in\mathbb{R}^{L\times d}$. Here $d$ is the shared projection dimension. We $\ell_2$-normalize $\vimg$, $\tcap$ and every row of $P$ and $H$ before scoring, writing $\operatorname{norm}(\bm{x})=\bm{x}/\lVert\bm{x}\rVert_2$.

As in LABCLIP~\citep{koishigarina2025clip}, we apply a learned linear transformation $M\in\mathbb{R}^{d\times d}$ to the sentence embedding before computing global cosine similarity:
\begin{equation}
    \bm{z}=\operatorname{norm}(M\tcap),
    \qquad
    s_{\mathrm{global}}(I,T)=\vimg^\top\bm{z}.
    \label{eq:global-score}
\end{equation}
Global-$M$ uses this score alone. Flow-CLIP jointly trains $M$, the token map $M_{\mathrm{tok}}$ and the residual map $W$, defined below. We initialize $M$ and $M_{\mathrm{tok}}$ as identity matrices and $W$ as zero, so the initial score equals the pretrained global cosine similarity.

\subsection{Image-Conditioned Evidence Refinement}

ICER constructs visual context from each token query's responses to the patches of the current image. Let $\mathcal{V}_T\subseteq\{1,\ldots,L\}$ contain the non-special, non-padding token positions, so $|\mathcal{V}_T|$ is the number of valid tokens. A learned matrix $M_{\mathrm{tok}}\in\mathbb{R}^{d\times d}$ defines the unit query
\begin{equation}
    \bm{q}_{\ell}=\operatorname{norm}(M_{\mathrm{tok}}\bm{h}_{\ell}).
    \label{eq:token-query}
\end{equation}
The valid-token set includes function words, punctuation and subwords. For token $\ell$ and patch $j$, let $a_{\ell j}=\bm{q}_{\ell}^{\top}\bm{p}_j$ denote their cosine similarity. We subtract the mean response across patches and retain its positive part, using above-mean responses to construct visual context:
\begin{equation}
\begin{aligned}
    e_{\ell j}
    &=
    \left[a_{\ell j}-\frac{1}{N}\sum_{k=1}^{N}a_{\ell k}\right]_{+},\\
    \bm{c}_{\ell}
    &=
    \frac{1}{N}\sum_{j=1}^{N}e_{\ell j}\bm{p}_{j}.
\end{aligned}
    \label{eq:centered-evidence}
\end{equation}
Here $[x]_+=\max(x,0)$, and $e_{\ell j}$ weights patch $j$ in the context vector $\bm{c}_{\ell}\in\mathbb{R}^{d}$. Uniform patch responses yield zero context, and centering makes the weights invariant to a common offset in patch similarity. The factor $1/N$ averages the weighted patch features without normalizing the weights to sum to one. Consequently, small deviations above the mean produce a small context vector, preserving the strength of the token's spatially varying response. For a fixed caption, the original token queries remain unchanged across images, while the patch weights and features determine the image-dependent input to the correction.

We project the visual context onto the tangent space at the unit token query before applying the learned residual map:
\begin{equation}
\begin{aligned}
    \bm{r}_{\ell}&=\bm{c}_{\ell}-(\bm{q}_{\ell}^{\top}\bm{c}_{\ell})\bm{q}_{\ell},\\
    \tilde{\bm{q}}_{\ell}&=\bm{q}_{\ell}+W\bm{r}_{\ell},\\
    \bm{q}^{\mathrm{res}}_{\ell}&=\operatorname{norm}(\tilde{\bm{q}}_{\ell}).
\end{aligned}
    \label{eq:residual-flow}
\end{equation}
The projection removes the component of $\bm{c}_{\ell}$ parallel to the unit query, giving $\bm{q}_{\ell}^{\top}\bm{r}_{\ell}=0$. The learned map $W\in\mathbb{R}^{d\times d}$ transforms this projected context into a residual update, yielding the intermediate query $\tilde{\bm{q}}_\ell$. Its normalization produces $\bm{q}^{\mathrm{res}}_{\ell}$ on the unit sphere, placing the original and corrected patch responses on the same cosine scale.

The local score measures the change in pooled patch evidence caused by query correction. For a unit query $\bm{q}$, define
\begin{equation}
    \mathcal{P}_{\tau_p}(\bm{q},P)
    =
    \tau_p\log\!\left[\frac{1}{N}\sum_{j=1}^{N}
    \exp(\bm{q}^{\top}\bm{p}_{j}/\tau_p)\right].
    \label{eq:patch-evidence}
\end{equation}
The temperature $\tau_p>0$ controls the smooth maximum over patch responses. We write $\rho_\ell=\mathcal{P}_{\tau_p}(\bm{q}_\ell,P)$ and $\rho_\ell^{\mathrm{res}}=\mathcal{P}_{\tau_p}(\bm{q}^{\mathrm{res}}_\ell,P)$ for pooled evidence before and after correction. Both queries are evaluated against the same patch set:
\begin{equation}
\begin{aligned}
    \Delta_{\ell}(I,T)
    &=\mathcal{P}_{\tau_p}(\bm{q}^{\mathrm{res}}_{\ell},P)
      -\mathcal{P}_{\tau_p}(\bm{q}_{\ell},P),\\
    s_{\mathrm{token}}(I,T)
    &=\frac{1}{|\mathcal{V}_T|}\sum_{\ell\in\mathcal{V}_T}\Delta_{\ell}(I,T).
\end{aligned}
    \label{eq:residual-increment}
\end{equation}
The sign of $\Delta_{\ell}$ indicates whether correction strengthens or weakens pooled support relative to the original query. Subtraction retains this distinction even when both queries have high absolute patch similarity. Token increments within a caption can have opposite signs. Averaging them over $\mathcal{V}_T$ yields the local term added to global compatibility in \Cref{eq:flowclip-score}. An empty valid-token set contributes zero; $W=\bm{0}$ or $\alpha=0$ recovers the global branch.

\subsection{Training Objective}\label{sec:training-objective}

Training jointly optimizes $M$, $M_{\mathrm{tok}}$ and $W$ through the image--caption score. A mini-batch $\mathcal{B}=\{(I_i,T_i^+,T_i^-,g_i)\}_{i=1}^{B}$ contains $B$ records, each with a matched image and caption, a full-caption hard negative, and image identity $g_i$. The indicator $y_{ij}=\mathbf{1}[g_i=g_j]$ marks positive captions belonging to the same image. With contrastive temperature $\tau_c>0$, define
\begin{equation}
    u_{ij}^{\pm}=\score(I_i,T_j^{\pm})/\tau_c.
    \label{eq:contrastive-logits}
\end{equation}
For image $I_i$, the image-to-text comparison uses logits $u_{ij}^{+}$ and $u_{ij}^{-}$ over all captions. For caption $T_i^+$, the text-to-image comparison uses $u_{ji}^{+}$ over all in-batch images. This asymmetry makes hard negatives compete with positive descriptions of an image, while retaining positive captions as the queries for image retrieval. The bidirectional multi-positive losses are
\begin{equation}
\begin{aligned}
    \mathcal{L}_{I2T}
    &=-\frac{1}{B}\sum_{i=1}^{B}
    \log
    \frac{\sum_{j=1}^{B}y_{ij}\exp(u_{ij}^{+})}
         {\sum_{j=1}^{B}\left[\exp(u_{ij}^{+})+\exp(u_{ij}^{-})\right]},
    \\
    \mathcal{L}_{T2I}
    &=-\frac{1}{B}\sum_{i=1}^{B}
    \log
    \frac{\sum_{j=1}^{B}y_{ij}\exp(u_{ji}^{+})}
         {\sum_{j=1}^{B}\exp(u_{ji}^{+})}.
\end{aligned}
\label{eq:training-objective}
\end{equation}
We denote the bidirectional contrastive objective with full-caption hard negatives by $\mathcal{L}_{\mathrm{HNB}}=\tfrac12(\mathcal{L}_{I2T}+\mathcal{L}_{T2I})$, abbreviated as HNB. Summing over shared image identities in the numerators accommodates multiple positive captions for an image.

CIMR compares the positive--negative caption margin across visual inputs. For each eligible record $i$, donor image $J_i$ has a different source identity and is used for both $T_i^+$ and $T_i^-$. Known positive matches are excluded from donor comparisons. Let $\mathcal{E}\subseteq\{1,\ldots,B\}$ contain records with an eligible hard-negative caption and donor, and define
\begin{equation}
    m_i(I)=\score(I,T_i^+)-\score(I,T_i^-).
    \label{eq:caption-margin}
\end{equation}
The Cross-Image Margin Ranking loss is
\begin{equation}
    \mathcal{L}_{\mathrm{CIMR}}
    =\frac{1}{|\mathcal{E}|}\sum_{i\in\mathcal{E}}
    \operatorname{softplus}\!\left(
      -\frac{m_i(I_i)-m_i(J_i)}{\tau_c}\right),
    \label{eq:cimr-objective}
\end{equation}
Here $\operatorname{softplus}(x)=\log(1+\exp x)$. Increasing $m_i(I_i)-m_i(J_i)$ lowers the loss, encouraging a larger caption margin on the matched image than on the donor. An additive text-only scoring term cancels because the same caption pair is used for both images. An additive image-only scoring term likewise cancels within each caption margin. CIMR therefore acts on score changes that depend jointly on image and caption content. For an empty eligible set, $\mathcal{L}_{\mathrm{CIMR}}=0$.

The complete training objective is
\begin{equation}
    \mathcal{L}=\mathcal{L}_{\mathrm{HNB}}
      +\lambda_{\mathrm{CIMR}}\mathcal{L}_{\mathrm{CIMR}}
    \label{eq:total-objective}
\end{equation}
We set $\lambda_{\mathrm{CIMR}}=0.1$. \Cref{alg:flowclip} summarizes the scoring procedure.

\section{Experiments}\label{sec:experiments}

\subsection{Experimental Setup}\label{sec:protocol}

\paragraph{\textbf{\upshape Training data.}}
We train on COCO~\citep{lin2014microsoft} train2014 with the Karpathy restval split~\citep{karpathy2015deep}, using 113,287 training images, 566,435 image--caption records and 25,000 validation records. Each training record pairs a human-written COCO caption~\citep{chen2015microsoft} with a full-caption hard negative. For the training-source comparison, we use Conceptual Captions (CC3M)~\citep{sharma2018conceptual} with the CC3M-100K protocol of FSC-CLIP~\citep{oh2024preserving}: 101,830 source records partitioned into 97,734 training and 4,096 validation records. CoCa~\citep{yu2022coca} supplies positive captions, and the negative-generation schedule uses phrase swaps, WordNet replacements and bigram shuffling. The released training and validation partitions share nine images, represented by 138 training and 16 validation records.

\paragraph{\textbf{\upshape Evaluation datasets and metrics.}}
We evaluate attribution, relations and caption order separately. ARO~\citep{yuksekgonul2023and} contains four subsets: VG-Attribution (4,517 images and 28,748 pairs), VG-Relation (5,316 images and 23,937 pairs), Flickr30k Order (1,000 images and 5,000 five-caption sets), and COCO Order (5,000 images and 25,010 five-caption sets). SugarCrepe~\citep{hsieh2023sugarcrepe} contains 7,511 fluent positive--negative pairs with Add, Replace and Swap edits. CREPE-Productivity~\citep{ma2023crepe} evaluates Atomic and Swap candidate sets at complexity levels 4--12.

Attribute--object binding evaluation uses ABC-6K~\citep{feng2023trainingfree} (3,213 comparison pairs), PUG:SPAR~\citep{bordes2023pug} (19,840 attribute-assignment pairs), and a two-object evaluation set derived from CLEVR~\citep{johnson2017clevr}. We abbreviate ABC-6K and PUG:SPAR as ABC and PUG in the result tables. The CLEVR combination-based partition contains 470 records, including 449 valid pairs with non-equivalent captions; the larger CLEVR evaluation pool contains 4,449 unambiguous pairs. The ablation table reports the 449-pair split, and results on the 4,449-pair pool are identified separately in the text. COCO-Spatial and GQA-Spatial contain 2,687 and 1,451 records, respectively, and follow the spatial-relation protocol of What's Up?~\citep{kamath2023whatsup}.

Retrieval evaluation uses Flickr30k~\citep{young2014image} and COCO~\citep{lin2014microsoft} with the Karpathy 1K and 5K splits~\citep{karpathy2015deep}, respectively: 1,000/5,000 images and 5,000/25,000 captions. Zero-shot classification uses CIFAR-10/CIFAR-100~\citep{krizhevsky2009learning} (10,000 test images each) and ImageNetV2~\citep{recht2019imagenetv2} (10,000 images with 1,000 class prompts). We retain each benchmark's task-level aggregation and specify the members of every cross-task mean in its caption. Pairwise accuracy tests the gold caption against one foil; candidate-set accuracy requires it to rank first among all alternatives. Retrieval extends this competition to the complete evaluation gallery, testing whether the learned score separates the correct match from many alternatives beyond the constructed foils. Image and comparison counts are reported separately because an image can appear in several evaluation records.


\paragraph{\textbf{\upshape Implementation details.}}
The main configuration freezes CLIP ViT-B/32~\citep{radford2021learning} and trains $M$, $M_{\mathrm{tok}}$ and $W$, totaling 786,432 parameters. For both head-only and encoder-adapted training, we use AdamW~\citep{loshchilov2019decoupled} with learning rate $8\times10^{-5}$, weight decay $10^{-4}$, $(\beta_1,\beta_2)=(0.9,0.999)$ and gradient clipping at a maximum norm of 1. Each batch contains 32 image--caption records and 32 full-caption negatives. COCO training uses five full epochs with a hard-negative pool of 501,921 eligible records. CC3M training uses three epochs, with 96,463, 97,394 and 97,734 valid records for phrase swaps, WordNet replacements and bigram shuffling, respectively.
We set $\tau_c=0.01$, $\alpha=1$ and $\lambda_{\mathrm{CIMR}}=0.1$, with $\tau_p=0.05$ for COCO and $0.06$ for CC3M. Experiments use NVIDIA RTX 3090 and A800 GPUs.
In the encoder-adapted variants, both the image and text encoders are trainable together with $M$, $M_{\mathrm{tok}}$ and $W$ under the HNB and CIMR objectives. Their COCO and CC3M results are evaluated after five and three training epochs, respectively. The scoring equations and objectives remain the same in both variants, allowing the comparison to assess the effect of updating encoder representations alongside the scoring maps.

\paragraph{\textbf{\upshape Feature extraction.}}
Scoring uses FP32 with autocast and TF32 disabled; normalized global and patch features are first rounded to FP16. CLIP~\citep{radford2021learning} inputs follow the OpenAI image preprocessing pipeline. Patch states pass through vision post-layer normalization and visual projection. Text token states use the text projection and the valid-token mask defined in the method. SigLIP~\citep{zhai2023siglip} retains its learned image-pooling operation and 196 patch embeddings, with a maximum text sequence length of 64 tokens in our evaluation. BLIP~\citep{li2022blip} uses 256-dimensional image--text contrastive (ITC) projections and 576 patch embeddings, with a maximum text sequence length of 35 tokens.

\subsection{Main Results}\label{sec:main-comparison}

\begin{table*}[pos=tbp]
	\centering
	\footnotesize
	\setlength{\tabcolsep}{2.0pt}
	
	\caption{Compositional matching and fine-grained binding accuracy (\%). ARO Avg. is the mean across four tasks; Swap-A/O denote SugarCrepe attribute/object swaps. \cmark/\xmark\ indicate frozen/jointly tuned encoders. Bold/underline indicate the best/second-best results within each training-source block.}
	\label{tab:clip-b32-coco-cc3m-comparison}
	\begin{adjustbox}{max width=\linewidth}
		\begin{tabular}{@{}llcccccccccccc@{}}
			\toprule
			\multirow{2}{*}[-2.5pt]{Method} & \multirow{2}{*}[-2.5pt]{Backbone} & \multirow{2}{*}[-2.5pt]{Training data} & \multirow{2}{*}[-2.5pt]{Frozen?} & \multicolumn{5}{c}{ARO} & \multicolumn{3}{c}{Attribute/Object Binding} & \multicolumn{2}{c}{Spatial Relations} \\
			\cmidrule(lr){5-9}\cmidrule(lr){10-12}\cmidrule(l){13-14}
			& & & & VG-A & VG-R & Flickr Ord. & COCO Ord. & Avg. & Swap-A & Swap-O & ABC & COCO-Sp. & GQA-Sp. \\
			\midrule
			CLIP~\citep{radford2021learning} & ViT-B/32 & -- & -- & 63.21 & 63.36 & 58.88 & 47.60 & 58.26 & 63.96 & 61.63 & 64.11 & 46.88 & 46.69 \\
			\midrule
			\multicolumn{14}{@{}l}{\textit{COCO-trained models}} \\
			NegCLIP~\citep{yuksekgonul2023and} & ViT-B/32 & COCO14-HN & \xmark & 70.50 & 81.12 & 90.50 & 86.97 & 82.27 & 75.23 & 75.51 & 73.58 & 46.54 & 45.91 \\
			LABCLIP~\citep{koishigarina2025clip} & ViT-B/32 & COCO14 & \cmark & 68.32 & 81.45 & 82.90 & 79.95 & 78.15 & 75.08 & 72.24 & 69.97 & \underline{48.24} & 45.18 \\
			FSC-CLIP~\citep{oh2024preserving} & ViT-B/32 & COCO-100K & \xmark & 70.67 & 80.53 & \textbf{95.50} & \textbf{94.08} & 85.19 & 78.23 & \textbf{76.73} & 74.76 & 47.91 & 46.17 \\
			DCSM~\citep{kang2025clip} & ViT-B/16 & COCO17-curated & \cmark & 56.59 & 61.80 & 30.40 & 32.63 & 45.36 & 59.76 & 53.06 & 61.53 & \textbf{72.47} & \textbf{68.45} \\
			CE-CLIP~\citep{zhang2024contrasting} & ViT-B/32 & COCO2014+2M & \xmark & \underline{76.30} & 83.86 & 79.18 & 80.91 & 80.06 & 77.03 & 73.06 & 73.36 & 47.50 & 47.38 \\
			\rowcolor{gray!12}
			Flow-CLIP & ViT-B/32 & COCO & \cmark & 74.10 & \underline{86.76} & \underline{92.12} & \underline{90.78} & \underline{85.94} & \underline{79.28} & \underline{75.92} & \underline{78.15} & 46.12 & 45.01 \\
			\rowcolor{gray!12}
			Flow-CLIP & ViT-B/32 & COCO & \xmark & \textbf{79.02} & \textbf{87.36} & 90.22 & 89.12 & \textbf{86.43} & \textbf{82.88} & \underline{75.92} & \textbf{81.92} & 48.04 & \underline{48.41} \\
			\midrule
			\multicolumn{14}{@{}l}{\textit{CC3M-trained models}} \\
			TSVLC~\citep{doveh2023teaching} & ViT-B/32 & CC3M & \xmark & \textbf{72.89} & 81.76 & 92.80 & 83.34 & 82.70 & 63.06 & 62.45 & 71.15 & 46.21 & 45.91 \\
			DAC-LLM~\citep{doveh2023dense} & ViT-B/32 & CC3M & \xmark & \underline{72.80} & 83.60 & 92.76 & 89.73 & \underline{84.72} & \underline{73.27} & \underline{71.43} & 70.59 & 45.38 & \underline{49.53} \\
			DAC-SAM~\citep{doveh2023dense} & ViT-B/32 & CC3M & \xmark & 69.15 & \underline{83.65} & \underline{94.72} & 88.50 & 84.00 & 71.77 & 70.20 & 70.21 & 47.65 & 48.66 \\
			LABCLIP~\citep{koishigarina2025clip} & ViT-B/32 & CC3M & \cmark & 68.01 & 78.45 & 83.16 & 72.73 & 75.59 & 67.42 & 68.16 & 67.88 & \underline{51.70} & 43.55 \\
			FSC-CLIP~\citep{oh2024preserving} & ViT-B/32 & CC3M-100K & \xmark & 69.49 & 80.47 & 90.98 & 90.40 & 82.83 & \textbf{75.53} & \textbf{73.88} & \underline{73.48} & 46.87 & 46.86 \\
			\rowcolor{gray!12}
			Flow-CLIP & ViT-B/32 & CC3M-100K & \cmark & 72.45 & 78.96 & 93.02 & \underline{91.63} & 84.01 & 70.57 & 67.76 & 72.49 & 47.35 & 47.76 \\
			\rowcolor{gray!12}
			Flow-CLIP & ViT-B/32 & CC3M-100K & \xmark & 70.66 & \textbf{83.99} & \textbf{96.96} & \textbf{96.25} & \textbf{86.96} & 71.17 & 70.20 & \textbf{74.95} & \textbf{52.93} & \textbf{49.87} \\
			\bottomrule
		\end{tabular}
	\end{adjustbox}
\end{table*}

The COCO-trained frozen Flow-CLIP scorer in \Cref{tab:clip-b32-coco-cc3m-comparison} reaches 78.15\% on ABC, exceeding NegCLIP~\citep{yuksekgonul2023and}, LABCLIP~\citep{koishigarina2025clip} and FSC-CLIP~\citep{oh2024preserving}, with a 3.39-percentage-point gain over FSC-CLIP. Because the ABC alternatives exchange attribute assignments while preserving the scene concepts, this gain reflects better discrimination between competing bindings. Its ARO average of 85.94\% combines attribution, relation and caption-order accuracy; the individual tasks show how these improvements are distributed.

The comparison with FSC-CLIP separates attribute-assignment gains from caption-order performance. Frozen Flow-CLIP scores higher on VG-Attribution, VG-Relation and Swap-Attribute, but lower on both caption-order tasks and Swap-Object. VG-Attribution, Swap-Attribute and ABC test which object has a named attribute, making their gains directly relevant to binding. The matched ablations in \Cref{sec:ablation} test the contribution of the local scoring architecture under a common training protocol.

With CC3M training, frozen Flow-CLIP exceeds CC3M-trained LABCLIP~\citep{koishigarina2025clip} on all four ARO tasks and on Swap-Attribute, ABC and GQA-Spatial, but scores lower on Swap-Object and COCO-Spatial. Its 84.01\% ARO average falls within the range of the encoder-tuned methods in the table, while both binding swap scores remain below FSC-CLIP~\citep{oh2024preserving}. These results extend the comparison to generated captions under the CC3M recipe and show that the binding advantage varies by task and comparator.

DCSM~\citep{kang2025clip} has the highest spatial scores in this comparison: 72.47\% on COCO-Spatial and 68.45\% on GQA-Spatial. It combines box-derived spatial descriptions, opposite arrangements generated with CutMix and functional rows for spatial words in the scoring map. Flow-CLIP learns from full-caption negatives and computes local support from image patches. The spatial comparison therefore includes differences in both supervision and scoring architecture.

Encoder adaptation yields ARO averages of 86.43\% with COCO and 86.96\% with CC3M, exceeding the frozen results in \Cref{tab:clip-b32-coco-cc3m-comparison} by 0.49 and 2.95 percentage points.

\begin{figure}[pos=tbp]
\centering
\includegraphics[width=\columnwidth]{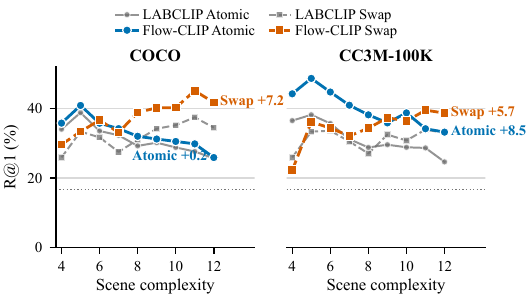}
\caption{CREPE Recall@1 for HNB-trained Flow-CLIP and LABCLIP~\citep{koishigarina2025clip} across complexity levels 4--12. Solid and dashed curves denote atomic and swap foils. Endpoint labels show gains over LABCLIP in percentage points at complexity 12.}
\label{fig:crepe-binding-gain}
\end{figure}

\paragraph{\textbf{\upshape Composition complexity.}}

\Cref{fig:crepe-binding-gain} compares Flow-CLIP and LABCLIP~\citep{koishigarina2025clip} on CREPE~\citep{ma2023crepe} atomic and swap foils across complexity levels 4--12. For the HNB-trained Flow-CLIP configurations shown, the gains over LABCLIP at complexity 12 are 0.2 and 7.2 Recall@1 percentage points on COCO atomic and swap foils, and 8.5 and 5.7 points on CC3M-100K. The COCO atomic curves converge at high complexity, while the swap curves retain a larger separation. On CC3M-100K, Flow-CLIP retains gains for both foil types at the highest tested complexity.

\subsection{Ablation Studies}\label{sec:ablation}

\begin{table*}[pos=tbp]
\centering
	\footnotesize
	\setlength{\tabcolsep}{3.0pt}
	\renewcommand{\arraystretch}{1.25}
	\definecolor{bindIncrease}{RGB}{0,86,125}
	\definecolor{bindDecrease}{RGB}{166,60,52}
	\newcommand{\binddelta}[2]{\textcolor{#1}{(#2)}}
	\caption{Component and objective ablation on COCO (\%). Checks indicate learned mappings and the CIMR loss. All trained rows use HNB. Lower lines show changes from CLIP~\citep{radford2021learning} in percentage points; Avg. is the mean across the 11 tasks. The gray configuration is complete Flow-CLIP.}
	\label{tab:residual-ablation}
	\begin{adjustbox}{max width=\textwidth}
		\begin{tabular}{@{}cccc@{\hspace{5pt}}*{12}{r}@{}}
			\toprule
			\multirow{2}{*}[-2.5pt]{$M$} & \multirow{2}{*}[-2.5pt]{$M_{\mathrm{tok}}$} & \multirow{2}{*}[-2.5pt]{$W$} & \multirow{2}{*}[-2.5pt]{CIMR} & \multicolumn{4}{c}{ARO} & \multicolumn{7}{c}{Fine-grained binding} & \multirow{2}{*}[-2.5pt]{Avg.} \\
			\cmidrule(lr){5-8}\cmidrule(lr){9-15}
			& & & & VG-A & VG-R & Flickr Ord. & COCO Ord. & Swap-A & Swap-O & ABC & COCO-Sp. & GQA-Sp. & PUG & CLEVR & \\
			\midrule
			-- & -- & -- & -- & 63.21 & 63.36 & 58.88 & 47.60 & 63.96 & 61.63 & 64.11 & 46.88 & 46.69 & 51.04 & 48.11 & 55.95 \\
			& & & & 71.04 & 84.23 & 91.24 & 90.20 & 76.43 & 72.24 & 71.12 & 49.91 & 43.03 & 50.63 & 44.99 & 67.73 \\
			\multirow{-2}{*}{\cmark} & \multirow{-2}{*}{--} & \multirow{-2}{*}{--} & \multirow{-2}{*}{--} & \binddelta{bindIncrease}{+7.83} & \binddelta{bindIncrease}{+20.88} & \binddelta{bindIncrease}{+32.36} & \binddelta{bindIncrease}{+42.61} & \binddelta{bindIncrease}{+12.46} & \binddelta{bindIncrease}{+10.61} & \binddelta{bindIncrease}{+7.00} & \binddelta{bindIncrease}{+3.03} & \binddelta{bindDecrease}{-3.65} & \binddelta{bindDecrease}{-0.41} & \binddelta{bindDecrease}{-3.12} & \binddelta{bindIncrease}{+11.78} \\
			& & & & 71.42 & 84.78 & 91.28 & 90.25 & 75.98 & 72.65 & 71.46 & 49.25 & 43.38 & 50.85 & 46.33 & 67.97 \\
			\multirow{-2}{*}{\cmark} & \multirow{-2}{*}{--} & \multirow{-2}{*}{--} & \multirow{-2}{*}{\cmark} & \binddelta{bindIncrease}{+8.21} & \binddelta{bindIncrease}{+21.43} & \binddelta{bindIncrease}{+32.40} & \binddelta{bindIncrease}{+42.65} & \binddelta{bindIncrease}{+12.01} & \binddelta{bindIncrease}{+11.02} & \binddelta{bindIncrease}{+7.35} & \binddelta{bindIncrease}{+2.37} & \binddelta{bindDecrease}{-3.31} & \binddelta{bindDecrease}{-0.19} & \binddelta{bindDecrease}{-1.78} & \binddelta{bindIncrease}{+12.01} \\
			& & & & 76.22 & 85.50 & 91.40 & 89.90 & 78.68 & 75.51 & 78.56 & 48.98 & 45.18 & 54.20 & 53.67 & 70.71 \\
			\multirow{-2}{*}{\cmark} & \multirow{-2}{*}{\cmark} & \multirow{-2}{*}{\cmark} & \multirow{-2}{*}{--} & \binddelta{bindIncrease}{+13.00} & \binddelta{bindIncrease}{+22.14} & \binddelta{bindIncrease}{+32.52} & \binddelta{bindIncrease}{+42.30} & \binddelta{bindIncrease}{+14.71} & \binddelta{bindIncrease}{+13.88} & \binddelta{bindIncrease}{+14.44} & \binddelta{bindIncrease}{+2.10} & \binddelta{bindDecrease}{-1.50} & \binddelta{bindIncrease}{+3.16} & \binddelta{bindIncrease}{+5.57} & \binddelta{bindIncrease}{+14.76} \\
			\rowcolor{gray!12}
			& & & & 76.03 & 86.16 & 91.42 & 90.04 & 77.93 & 77.55 & 78.34 & 48.53 & 44.97 & 55.05 & 56.12 & 71.10 \\
			\rowcolor{gray!12}
			\multirow{-2}{*}{\cmark} & \multirow{-2}{*}{\cmark} & \multirow{-2}{*}{\cmark} & \multirow{-2}{*}{\cmark} & \binddelta{bindIncrease}{+12.82} & \binddelta{bindIncrease}{+22.80} & \binddelta{bindIncrease}{+32.54} & \binddelta{bindIncrease}{+42.44} & \binddelta{bindIncrease}{+13.96} & \binddelta{bindIncrease}{+15.92} & \binddelta{bindIncrease}{+14.22} & \binddelta{bindIncrease}{+1.65} & \binddelta{bindDecrease}{-1.72} & \binddelta{bindIncrease}{+4.01} & \binddelta{bindIncrease}{+8.02} & \binddelta{bindIncrease}{+15.15} \\
			\bottomrule
		\end{tabular}
	\end{adjustbox}
\end{table*}

We independently train Global-$M$ and the global/local scorers with the same data and optimization settings, evaluating all heads at COCO epoch 5. Means and standard deviations across three runs quantify training variability.

\paragraph{\textbf{\upshape Component effects.}}

Global calibration accounts for much of the improvement over pretrained CLIP~\citep{radford2021learning} in \Cref{tab:residual-ablation}. With matched COCO data and optimization at epoch 5, Global-$M$ raises the 11-task mean from 55.95\% to 67.73\%. Its largest gains occur on caption order, with accuracy above 90\% on both Flickr30k Order and COCO Order. Flow-CLIP reaches 71.10\%, adding 3.37 percentage points to the mean. The binding tasks distinguish this additional benefit from the caption-order gains already obtained through global calibration.

The seven-task binding mean covers Swap-A, Swap-O, ABC, COCO-Spatial, GQA-Spatial, PUG and the CLEVR 449-pair split. It rises from 54.63\% for CLIP to 58.34\% for Global-$M$ and 62.64\% for Flow-CLIP: global calibration adds 3.70 percentage points, and the joint scorer adds another 4.30 points. This group tests attribute assignments, object identity and spatial relations. The 11-task mean additionally includes all four ARO tasks, where global calibration already achieves high caption-order accuracy.

For a fixed global mapping, the local contribution in \Cref{eq:residual-increment} changes the caption-pair margin by $\alpha[s_{\mathrm{token}}(I,T^+)-s_{\mathrm{token}}(I,T^-)]$. Ranking therefore depends on the difference between the two evidence changes. Equal increments leave the margin unchanged; unequal increments can increase or decrease it. In the trained ablations, each scorer learns its own global mapping, jointly with the local maps where present. The HNB and CIMR comparisons measure the effect of training the global/local architecture under each objective.

Controlled assignment tasks preserve the objects and attributes mentioned in the captions while changing which attribute belongs to each object. Under HNB, training the local branch with the global map raises accuracy on the CLEVR~\citep{johnson2017clevr} 449-pair split from 44.99\% to 53.67\% and on PUG:SPAR~\citep{bordes2023pug} from 50.63\% to 54.20\%. With CIMR in both heads, Flow-CLIP exceeds Global-$M$+CIMR by 9.80 and 4.20 percentage points, respectively. Training with the local branch improves both assignment tasks under either objective. On the CLEVR 4,449-pair evaluation pool, HNB-trained Global-$M$ achieves 51.29\%, while Flow-CLIP trained with HNB and CIMR achieves 56.06\%.

Flow-CLIP exceeds Global-$M$ by 5.31 percentage points on Swap-Object and 7.22 points on ABC, while COCO-Spatial~\citep{kamath2023whatsup} decreases by 1.38 points. The local architecture's gains on binding thus differ from its effect on spatial-relation discrimination. Paired predictions distinguish corrected selections from new errors.

\paragraph{\textbf{\upshape Encoder adaptation.}}

The encoder-update ablation compares the fully trainable Flow-CLIP models in \Cref{tab:clip-b32-coco-cc3m-comparison} with separate frozen Flow-CLIP runs on the corresponding training source. The frozen controls achieve 85.91\% on COCO and 83.85\% on CC3M, giving adaptation gains of 0.52 and 3.11 points, respectively. Within the matched COCO comparison, Flickr30k Order, COCO Order, Swap-Object and COCO-Spatial decrease by 1.20, 0.91, 1.63 and 0.49 points. The ARO increase comes from attribution and relation gains despite lower caption-order accuracy.

\paragraph{\textbf{\upshape Training variability.}}
Across three independently trained COCO runs, mean ARO accuracy and its sample standard deviation are $84.31\pm0.19$ for Global-$M$, $84.47\pm0.21$ for Global-$M$+CIMR, $85.77\pm0.25$ for Flow-CLIP (HNB) and $85.90\pm0.04$ for Flow-CLIP with CIMR. Flow-CLIP exceeds Global-$M$ in every run.

\paragraph{\textbf{\upshape Confidence intervals.}}
We estimate 95\% percentile confidence intervals for three-run mean accuracy with 10,000 paired image-cluster bootstrap draws at fixed final checkpoints. Each draw uses the same image resampling weights across methods, tasks and training runs. The intervals for the Flow-CLIP minus Global-$M$ accuracy difference are [1.26, 2.31] percentage points on SugarCrepe, [5.95, 8.13] on ABC and [$-3.20$, $-0.04$] on COCO-Spatial. They support a gain on SugarCrepe and ABC and a decline on COCO-Spatial under this evaluation protocol.

Standard deviations across runs measure training variability, while the paired bootstrap estimates image-sampling uncertainty in the three-run mean at fixed checkpoints. Each bootstrap draw must retain all benchmark aggregation groups. Some VG-Relation groups contain only one unique image, so we omit intervals for VG-Relation and the ARO mean under this protocol. We also omit GQA and image-replacement intervals because image-cluster membership, including shared donor groups, cannot be fully determined from the evaluation records.

\paragraph{\textbf{\upshape Decision changes.}}

Paired predictions show which Global-$M$ errors Flow-CLIP corrects and which correct predictions it loses. These transitions expose changes hidden by average accuracy when repairs and regressions offset each other. On 7,511 SugarCrepe~\citep{hsieh2023sugarcrepe} records, Flow-CLIP repairs 366 errors and introduces 231, yielding 135 additional correct decisions. On the 3,213 ABC~\citep{feng2023trainingfree} pairs, it repairs 352 and introduces 120 errors. In both comparisons, repairs exceed regressions on identical images and candidate captions, with a larger net gain on ABC's attribute-assignment task.

COCO-Spatial shows the reverse balance: 155 repairs and 213 regressions. Its official score decreases by 1.38 percentage points with equal weighting of the one- and two-object subsets. This weighting differs from pooling correct decisions across all records, so the score difference and net repair count use different aggregations. Here, training the joint scorer introduces more spatial errors than it corrects.

Among records correctly ranked by Global-$M$, Flow-CLIP produces smaller gold--foil margins on 68.06\% of COCO-Spatial and 67.66\% of GQA-Spatial records across the three runs, using the benchmarks' subset weights. A smaller margin can remain positive, leaving the pairwise ranking correct.

\begin{table}[pos=tbp]
	\centering
	\footnotesize
	\setlength{\tabcolsep}{2.3pt}
	\caption{ARO accuracy across CIMR weight $\lambda_{\mathrm{CIMR}}$, pooling temperature $\tau_p$ and local scale $\alpha$ (\%). Avg. is the mean across four tasks; the gray row is the reference configuration.}
	\label{tab:hyperparameter-sensitivity}
	\begin{adjustbox}{max width=\linewidth}
		\begin{tabular}{@{}rrrrrrrr@{}}
			\toprule
			$\lambda_{\mathrm{CIMR}}$ & $\tau_p$ & $\alpha$ & VG-A & VG-R & Flickr Ord. & COCO Ord. & Avg. \\
			\midrule
			0.05 & 0.03 & 1.00 & 76.00 & 86.40 & 91.54 & \underline{90.14} & 86.02 \\
			0.05 & 0.05 & 1.00 & \textbf{76.06} & 86.02 & 91.38 & 90.04 & 85.88 \\
			0.05 & 0.03 & 0.75 & 75.75 & 86.37 & 91.48 & 90.09 & 85.92 \\
			0.05 & 0.05 & 0.75 & 76.03 & 86.16 & 91.42 & 90.04 & 85.91 \\
			\midrule
			0.10 & 0.03 & 1.00 & 75.65 & \textbf{86.75} & 91.54 & 90.08 & 86.00 \\
			\rowcolor{gray!15}
			0.10 & 0.05 & 1.00 & \underline{76.05} & 86.45 & \underline{91.78} & \textbf{90.22} & \textbf{86.12} \\
			0.10 & 0.03 & 0.75 & 75.61 & 86.54 & \textbf{91.90} & 90.04 & 86.02 \\
			0.10 & 0.05 & 0.75 & 75.84 & \underline{86.67} & 91.74 & 90.13 & \underline{86.09} \\
			\bottomrule
		\end{tabular}
	\end{adjustbox}
\end{table}

\paragraph{\textbf{\upshape Hyperparameter sensitivity.}}
\label{sec:sensitivity}

Mean ARO accuracy ranges from 85.88\% to 86.12\% across the eight configurations in \Cref{tab:hyperparameter-sensitivity}. We vary $\lambda_{\mathrm{CIMR}}\in\{0.05,0.10\}$, $\tau_p\in\{0.03,0.05\}$ and $\alpha\in\{0.75,1.00\}$ using frozen CLIP~\citep{radford2021learning} ViT-B/32 and five COCO epochs. The default is $(0.10,0.05,1.00)$.

The effect of pooling temperature depends on the CIMR weight. Increasing $\tau_p$ from 0.03 to 0.05 lowers the ARO mean at $\lambda_{\mathrm{CIMR}}=0.05$ but raises it at $\lambda_{\mathrm{CIMR}}=0.10$ for both local scales. Temperature controls emphasis on the highest patch responses, while $\alpha$ scales the signed local correction.

Similar ARO means mask opposing changes in the constituent tasks: increasing the CIMR weight raises VG-Relation and lowers VG-Attribution in all four matched $(\tau_p,\alpha)$ pairs. Increasing $\alpha$ at fixed CIMR weight and pooling temperature raises VG-Attribution in each tested comparison and leaves COCO Order unchanged or improves it. Flickr30k Order changes in both directions.

\subsection{Image Replacement}\label{sec:image-replacement}

\begin{table*}[pos=tbp]
\centering
\footnotesize
\setlength{\tabcolsep}{3pt}
\renewcommand{\arraystretch}{1.12}
\caption{Image-replacement controls for compositional matching (\%). FSC-CLIP and the four matched scoring controls each use one checkpoint across all three conditions. Real uses matched images; Shuffle uses a shared permutation; Fixed uses one anchor. The 6-task average covers SugarCrepe Add/Replace/Swap, ABC, COCO-Spatial and GQA-Spatial.}
\label{tab:prior-controls}
\begin{adjustbox}{max width=\textwidth}
\begin{tabular}{@{}ll*{12}{r}@{}}
\toprule
\multirow{2}{*}[-2.5pt]{Method} & \multirow{2}{*}[-2.5pt]{Image control} & \multicolumn{5}{c}{ARO} & \multicolumn{3}{c}{SugarCrepe} & \multirow{2}{*}[-2.5pt]{ABC} & \multicolumn{2}{c}{Spatial} & \multicolumn{1}{c}{6-task} \\
\cmidrule(lr){3-7}\cmidrule(lr){8-10}\cmidrule(lr){12-13}\cmidrule(l){14-14}
& & VG-A & VG-R & Flickr Ord. & COCO Ord. & Avg. & Add & Repl. & Swap & & COCO-Sp. & GQA-Sp. & Avg. \\
\midrule
\multicolumn{14}{@{}l}{\textit{Reference methods}} \\
 & Real & 63.21 & 63.36 & 58.88 & 47.60 & 58.26 & 75.09 & 80.76 & 63.34 & 64.11 & 46.88 & 46.69 & 62.81 \\
 & Shuffle & 47.66 & 41.11 & 10.36 & 7.52 & 26.66 & 62.27 & 46.72 & 43.25 & 44.69 & 46.45 & 50.17 & 48.93 \\
\multirow{-3}{*}{CLIP~\citep{radford2021learning}} & Fixed & 48.63 & 51.87 & 5.54 & 8.33 & 28.59 & 55.16 & 46.57 & 44.02 & 43.67 & 46.01 & 50.82 & 47.71 \\
\midrule
 & Real & 68.32 & 81.45 & 82.90 & 79.95 & 78.15 & 81.02 & 83.25 & 73.66 & 69.97 & 48.24 & 45.18 & 66.89 \\
 & Shuffle & 54.35 & 64.92 & 47.30 & 34.50 & 50.27 & 54.33 & 48.35 & 53.16 & 50.58 & 50.21 & 48.84 & 50.91 \\
\multirow{-3}{*}{LABCLIP~\citep{koishigarina2025clip}} & Fixed & 55.05 & 66.15 & 35.42 & 49.32 & 51.49 & 54.38 & 45.08 & 52.43 & 50.17 & 43.61 & 48.28 & 48.99 \\
\midrule
 & Real & 70.50 & 81.12 & 90.50 & 86.97 & 82.27 & 85.80 & 85.02 & 75.37 & 73.58 & 46.54 & 45.91 & 68.70 \\
 & Shuffle & 52.30 & 54.69 & 29.66 & 20.45 & 39.28 & 76.30 & 50.04 & 52.64 & 50.20 & 48.12 & 48.63 & 54.32 \\
\multirow{-3}{*}{NegCLIP~\citep{yuksekgonul2023and}} & Fixed & 56.26 & 66.38 & 17.70 & 33.51 & 43.46 & 69.79 & 47.09 & 48.95 & 49.02 & 47.50 & 49.40 & 51.96 \\
\midrule
 & Real & 70.67 & 80.53 & 95.50 & 94.08 & 85.19 & 88.98 & 87.32 & 77.48 & 74.76 & 47.91 & 46.17 & 70.44 \\
 & Shuffle & 47.47 & 34.20 & 23.20 & 14.48 & 29.84 & 77.82 & 49.34 & 45.50 & 45.04 & 48.97 & 49.23 & 52.65 \\
\multirow{-3}{*}{FSC-CLIP~\citep{oh2024preserving}} & Fixed & 50.12 & 45.79 & 7.26 & 17.95 & 30.28 & 71.48 & 45.24 & 41.65 & 44.26 & 46.95 & 49.01 & 49.77 \\
\midrule
\multicolumn{14}{@{}l}{\textit{Matched scoring controls}} \\
 & Real & 71.04 & 84.23 & 91.24 & 90.20 & 84.18 & 79.23 & 82.01 & 75.30 & 71.12 & 49.91 & 43.03 & 66.77 \\
 & Shuffle & 57.47 & 64.16 & 48.96 & 38.22 & 52.20 & 58.17 & 47.50 & 54.99 & 52.32 & 49.51 & 49.96 & 52.08 \\
\multirow{-3}{*}{Global-$M$} & Fixed & 58.19 & 71.94 & 43.82 & 49.76 & 55.93 & 50.87 & 44.62 & 50.71 & 50.39 & 53.15 & 48.75 & 49.75 \\
\midrule
 & Real & 71.42 & 84.78 & 91.28 & 90.25 & 84.43 & 79.96 & 82.16 & 75.08 & 71.46 & 49.25 & 43.38 & 66.88 \\
 & Shuffle & 53.45 & 52.80 & 25.28 & 14.12 & 36.41 & 58.90 & 48.47 & 46.43 & 48.43 & 50.62 & 48.37 & 50.20 \\
\multirow{-3}{*}{Global-$M$+CIMR} & Fixed & 53.37 & 63.66 & 19.20 & 21.06 & 39.32 & 50.80 & 44.93 & 44.57 & 46.69 & 50.45 & 48.54 & 47.66 \\
\midrule
 & Real & 76.22 & 85.50 & 91.40 & 89.90 & 85.75 & 80.79 & 83.02 & 77.83 & 78.56 & 48.98 & 45.18 & 69.06 \\
 & Shuffle & 57.52 & 67.22 & 40.72 & 29.52 & 48.75 & 61.11 & 46.96 & 53.90 & 51.45 & 47.88 & 49.27 & 51.76 \\
\multirow{-3}{*}{Flow-CLIP (HNB)} & Fixed & 56.94 & 70.47 & 36.60 & 36.67 & 50.17 & 55.85 & 43.11 & 51.92 & 50.82 & 46.82 & 49.36 & 49.65 \\
\midrule
\rowcolor{gray!12}
 & Real & 76.03 & 86.16 & 91.42 & 90.04 & 85.91 & 81.66 & 83.18 & 77.83 & 78.34 & 48.53 & 44.97 & 69.08 \\
\rowcolor{gray!12}
 & Shuffle & 52.57 & 58.21 & 21.04 & 10.83 & 35.66 & 59.22 & 46.67 & 47.09 & 49.49 & 50.12 & 49.66 & 50.38 \\
\rowcolor{gray!12}
\multirow{-3}{*}{Flow-CLIP} & Fixed & 51.75 & 65.78 & 15.82 & 15.50 & 37.21 & 55.08 & 43.40 & 46.21 & 46.03 & 46.48 & 49.61 & 47.80 \\
\bottomrule
\end{tabular}
\end{adjustbox}
\end{table*}

The controls in \Cref{tab:prior-controls} change the visual input while keeping caption candidates fixed. \textit{Real} uses the matched image and measures accuracy with the intended visual evidence. \textit{Shuffle} permutes image assignments, giving all candidates for a record the same replacement image. It disrupts image--caption correspondence while retaining variation in visual input across records. \textit{Fixed} uses one anchor image for all records within a task, removing this variation as well. The two replacement conditions test whether preference for the original gold caption persists across different replacement images or under one constant visual input.

FSC-CLIP~\citep{oh2024preserving} uses its COCO-trained checkpoint across all three image conditions. Each of the four matched scoring controls likewise uses the same checkpoint across conditions. The HNB and CIMR comparisons within each architecture therefore measure how the training objective affects responses to image replacement.

Accuracy after replacement measures how often a scorer retains the gold-caption preference with a different image. Gold labels remain tied to the original images; interpreting this retention therefore requires the corresponding matched-image accuracy.

HNB-trained Flow-CLIP achieves mean ARO accuracy of 85.75\% on matched images, 48.75\% under Shuffle and 50.17\% under Fixed. Adding CIMR during training yields 85.91\% on matched images and lowers Shuffle and Fixed accuracy to 35.66\% and 37.21\%, decreases of 13.08 and 12.96 percentage points. Adding CIMR to Global-$M$ also lowers ARO accuracy under image replacement, with little change in matched-image accuracy. For both architectures, caption-order tasks show larger reductions under Shuffle and Fixed than attribution or relation tasks.

The caption candidates and their linguistic cues remain unchanged across these conditions. CIMR thus mitigates reliance on language priors in compositional matching, reducing the gold-caption preference retained under image replacement while maintaining matched-image accuracy.

For Flow-CLIP with the CIMR objective in \Cref{eq:cimr-objective}, image replacement produces a larger accuracy drop on ARO than on the six-task average over SugarCrepe, ABC and spatial-relation tasks, which remains 50.38\% under Shuffle.

The matched-image ablation in \Cref{tab:residual-ablation} shows that adding CIMR to the HNB-trained global/local scorer raises CLEVR and PUG accuracy by 2.45 and 0.85 percentage points, while Swap-Attribute, ABC, COCO-Spatial and GQA-Spatial decrease by 0.75, 0.22, 0.45 and 0.21 points.

Lower caption-preference retention under image replacement coexists with gains and losses across the matched-image binding tasks.

\subsection{Encoder Comparisons}\label{sec:encoder-generalization}

\begin{table*}[pos=tb]
\centering
\footnotesize
\setlength{\tabcolsep}{1.5pt}
\renewcommand{\arraystretch}{1.12}
\caption{Frozen encoder families with COCO-trained scoring heads (\%). Base denotes pretrained scoring. The 5-task binding average covers Swap-A, Swap-O, ABC, COCO-Spatial and GQA-Spatial; Sugar Avg. is the mean across Add, Replace and Swap. T2I/I2T report COCO Recall@1 in the two retrieval directions. Bold/underline indicate the best/second-best scores within each backbone.}
\label{tab:cross-family-transfer}
\label{tab:cross-family-full}
\begin{adjustbox}{max width=\textwidth}
\begin{tabular}{@{}ll*{17}{r}@{}}
\toprule
\multirow{2}{*}[-6pt]{Backbone} & \multirow{2}{*}[-6pt]{Scorer} & \multicolumn{5}{c}{ARO} & \multicolumn{6}{c}{Fine-grained binding} & \multicolumn{4}{c}{SugarCrepe} & \multicolumn{2}{c}{COCO retrieval} \\
\cmidrule(lr){3-7}\cmidrule(lr){8-13}\cmidrule(lr){14-17}\cmidrule(l){18-19}
& & VG-A & VG-R & \shortstack{Flickr\\Ord.} & \shortstack{COCO\\Ord.} & Avg. & \shortstack{Swap-\\A} & \shortstack{Swap-\\O} & ABC & \shortstack{COCO-\\Sp.} & \shortstack{GQA-\\Sp.} & \shortstack{5-task\\Avg.} & Add & Repl. & Swap & \shortstack{Sugar\\Avg.} & T2I & I2T \\
\midrule
\multirow{3}{*}{CLIP ViT-B/32~\citep{radford2021learning}} & Base & 63.21 & 63.36 & 58.88 & 47.60 & 58.26 & 63.96 & 61.63 & 64.11 & 46.88 & \textbf{46.69} & 56.66 & 75.09 & 80.76 & 63.34 & 73.06 & 30.22 & \textbf{50.44} \\
 & Global-$M$ & \underline{71.04} & \underline{84.23} & \underline{91.24} & \textbf{90.20} & \underline{84.18} & \underline{76.43} & \underline{72.24} & \underline{71.12} & \textbf{49.91} & 43.03 & \underline{62.55} & \underline{79.23} & \underline{82.01} & \underline{75.30} & \underline{78.85} & \underline{38.06} & 49.60 \\
\rowcolor{gray!12}
 & Flow-CLIP & \textbf{76.03} & \textbf{86.16} & \textbf{91.42} & \underline{90.04} & \textbf{85.91} & \textbf{77.93} & \textbf{77.55} & \textbf{78.34} & \underline{48.53} & \underline{44.97} & \textbf{65.46} & \textbf{81.66} & \textbf{83.18} & \textbf{77.83} & \textbf{80.89} & \textbf{38.28} & \underline{50.14} \\
\midrule
\multirow{3}{*}{SigLIP ViT-B/16~\citep{zhai2023siglip}} & Base & 55.87 & 32.04 & 41.44 & 33.07 & 40.60 & \underline{71.47} & 60.41 & 72.49 & 47.55 & \underline{48.36} & 60.06 & 87.69 & 84.56 & 68.50 & 80.25 & 47.84 & \textbf{65.76} \\
 & Global-$M$ & \underline{78.77} & \underline{86.52} & \underline{95.50} & \underline{94.87} & \underline{88.91} & \textbf{85.74} & \textbf{74.69} & \underline{81.57} & \underline{50.76} & \textbf{48.75} & \textbf{68.30} & \underline{88.05} & \underline{87.62} & \textbf{82.77} & \underline{86.15} & \textbf{49.50} & 60.50 \\
\rowcolor{gray!12}
 & Flow-CLIP & \textbf{80.77} & \textbf{86.99} & \textbf{95.98} & \textbf{95.94} & \textbf{89.92} & \textbf{85.74} & \underline{73.47} & \textbf{81.95} & \textbf{51.44} & 47.55 & \underline{68.03} & \textbf{88.38} & \textbf{88.14} & \underline{82.44} & \textbf{86.32} & \underline{48.97} & \underline{61.58} \\
\midrule
\multirow{3}{*}{BLIP ITC-Base~\citep{li2022blip}} & Base & 81.83 & 53.17 & 28.22 & 21.80 & 46.26 & 84.38 & 75.10 & 86.49 & 45.72 & \textbf{49.14} & 68.17 & \textbf{95.53} & 90.54 & 81.89 & 89.32 & \textbf{57.08} & \textbf{73.98} \\
 & Global-$M$ & \textbf{86.31} & \textbf{88.97} & \textbf{94.76} & \underline{95.97} & \textbf{91.50} & \underline{87.99} & \underline{79.59} & \underline{88.45} & \underline{49.40} & \underline{47.38} & \underline{70.56} & \underline{93.03} & \underline{92.80} & \underline{85.73} & \underline{90.52} & \underline{56.99} & \underline{71.02} \\
\rowcolor{gray!12}
 & Flow-CLIP & \underline{83.98} & \underline{88.62} & \underline{94.64} & \textbf{96.17} & \underline{90.85} & \textbf{89.04} & \textbf{80.41} & \textbf{88.80} & \textbf{50.10} & 47.16 & \textbf{71.10} & 92.92 & \textbf{92.93} & \textbf{86.72} & \textbf{90.86} & 56.83 & 69.88 \\
\bottomrule
\end{tabular}
\end{adjustbox}
\end{table*}

\begin{table*}[pos=tbp]
\centering
\footnotesize
\setlength{\tabcolsep}{3pt}
\renewcommand{\arraystretch}{1.12}
\caption{Frozen CLIP~\citep{radford2021learning} backbones with COCO-trained scoring heads (\%). ARO Avg. is the mean across four tasks; the 5-task binding average covers Swap-A, Swap-O, ABC, COCO-Spatial and GQA-Spatial; SugarCrepe Avg. is the mean across Add, Replace and Swap. T2I/I2T denote COCO text-to-image/image-to-text Recall@1. Bold/underline indicate the best/second-best scores within each backbone.}
\label{tab:clip-scale}
\begin{adjustbox}{max width=\textwidth}
\begin{tabular}{@{}ll*{9}{r}@{}}
\toprule
\multirow{2}{*}[-2.5pt]{Backbone} & \multirow{2}{*}[-2.5pt]{Scorer} & \multicolumn{5}{c}{ARO} & \multicolumn{1}{c}{Binding (5-task)} & \multicolumn{1}{c}{SugarCrepe} & \multicolumn{2}{c}{COCO retrieval} \\
\cmidrule(lr){3-7}\cmidrule(lr){8-8}\cmidrule(lr){9-9}\cmidrule(l){10-11}
& & VG-A & VG-R & Flickr Ord. & COCO Ord. & Avg. & Avg. & Avg. & T2I & I2T \\
\midrule
\multirow{4}{*}{ViT-B/32~\citep{radford2021learning}} & Base & 63.21 & 63.36 & 58.88 & 47.60 & 58.26 & 56.66 & 73.06 & 30.22 & \textbf{50.44} \\
 & Global-$M$ & 71.04 & 84.23 & 91.24 & \underline{90.20} & 84.18 & \underline{62.55} & 78.85 & 38.06 & 49.60 \\
 & Global-$M$+CIMR & \underline{71.42} & \underline{84.78} & \underline{91.28} & \textbf{90.25} & \underline{84.43} & 62.54 & \underline{79.07} & \underline{38.16} & 48.68 \\
\rowcolor{gray!12}
 & Flow-CLIP & \textbf{76.03} & \textbf{86.16} & \textbf{91.42} & 90.04 & \textbf{85.91} & \textbf{65.46} & \textbf{80.89} & \textbf{38.28} & \underline{50.14} \\
\midrule
\multirow{4}{*}{ViT-B/16~\citep{radford2021learning}} & Base & 62.56 & 63.91 & 57.06 & 49.78 & 58.33 & 57.70 & 73.39 & 32.97 & 52.46 \\
 & Global-$M$ & \underline{71.87} & 83.86 & \textbf{92.56} & 92.10 & 85.10 & \underline{64.45} & \underline{79.62} & \underline{41.28} & \underline{53.30} \\
 & Global-$M$+CIMR & 71.53 & \underline{84.27} & \underline{92.48} & \textbf{92.25} & \underline{85.13} & 63.98 & 79.49 & \textbf{41.33} & 51.94 \\
\rowcolor{gray!12}
 & Flow-CLIP & \textbf{73.34} & \textbf{85.13} & \underline{92.48} & \underline{92.19} & \textbf{85.78} & \textbf{65.57} & \textbf{81.53} & 41.26 & \textbf{53.54} \\
\midrule
\multirow{4}{*}{ViT-L/14~\citep{radford2021learning}} & Base & 61.87 & 64.63 & 55.80 & 46.56 & 57.22 & 56.29 & 72.97 & 36.14 & \underline{56.20} \\
 & Global-$M$ & \underline{69.38} & \textbf{84.67} & \underline{93.70} & \underline{93.21} & \underline{85.24} & \underline{64.79} & \underline{79.84} & \underline{43.51} & 55.44 \\
 & Global-$M$+CIMR & 68.74 & \underline{84.12} & \textbf{94.02} & \textbf{93.27} & 85.04 & 64.23 & 79.55 & 43.46 & 54.28 \\
\rowcolor{gray!12}
 & Flow-CLIP & \textbf{73.61} & 83.81 & 93.26 & 92.90 & \textbf{85.90} & \textbf{67.09} & \textbf{82.99} & \textbf{45.19} & \textbf{57.26} \\
\bottomrule
\end{tabular}
\end{adjustbox}
\end{table*}

Flow-CLIP's gains occur in different task groups for SigLIP~\citep{zhai2023siglip} and BLIP~\citep{li2022blip} in \Cref{tab:cross-family-transfer}. Both learned heads improve ARO over the unadapted encoders. Relative to Global-$M$, Flow-CLIP increases SigLIP's ARO mean from 88.91\% to 89.92\%, while binding decreases from 68.30\% to 68.03\%. BLIP exhibits the opposite pattern: ARO decreases from 91.50\% to 90.85\%, while binding increases from 70.56\% to 71.10\%.

All heads are trained on COCO using their frozen encoder's features, which differ in image pooling, patch resolution and projection dimension. Relative to unadapted BLIP, both learned heads improve ARO accuracy but reduce COCO image-to-text Recall@1.

Binding gains over Global-$M$ persist across the three CLIP~\citep{radford2021learning} backbones in \Cref{tab:clip-scale}, but their size does not increase monotonically with backbone capacity. For ViT-B/32, ViT-B/16 and ViT-L/14, the five-task mean increases by 2.91, 1.12 and 2.30 percentage points, respectively, while Flow-CLIP's ARO averages range from 85.78\% to 85.91\%. On ViT-L/14, the binding and SugarCrepe means improve alongside both retrieval directions, despite decreases in several ARO tasks.

\subsection{Classification and Retrieval}\label{sec:general-tasks}

\begin{table}[pos=!tbp]
\centering
\footnotesize
\setlength{\tabcolsep}{1.25pt}
\renewcommand{\arraystretch}{1.14}
\caption{Zero-shot Top-1 classification and Recall@1 retrieval (\%). T2I retrieves images from captions; I2T retrieves captions from images. Colored parentheses report percentage-point changes from frozen CLIP (blue: gain; red: loss).}
\label{tab:zero-shot-preservation}
\begin{tabular*}{\columnwidth}{@{\extracolsep{\fill}}lrrrrrr@{}}
\toprule
\multirow{2}{*}[-2.5pt]{Dataset} & \multirow{2}{*}[-2.5pt]{CLIP} & \multirow{2}{*}[-2.5pt]{NegCLIP} & \multicolumn{2}{c}{Global-$M$} & \multicolumn{2}{c}{Flow-CLIP} \\
\cmidrule(lr){4-5}\cmidrule(l){6-7}
& & & COCO & CC3M & COCO & CC3M \\
\midrule
\multicolumn{7}{@{}l}{\textit{Classification (Top-1)}} \\
CIFAR-10 & 88.3 & 89.0 & 86.1 & 85.2 & 86.5\,{\fontsize{6}{7.2}\selectfont\textcolor{red!70!black}{$(-1.8)$}} & 86.4\,{\fontsize{6}{7.2}\selectfont\textcolor{red!70!black}{$(-1.9)$}} \\
CIFAR-100 & 64.5 & 63.0 & 58.9 & 58.9 & 58.0\,{\fontsize{6}{7.2}\selectfont\textcolor{red!70!black}{$(-6.5)$}} & 59.3\,{\fontsize{6}{7.2}\selectfont\textcolor{red!70!black}{$(-5.2)$}} \\
ImageNetV2 & 54.7 & 48.7 & 45.3 & 50.1 & 47.0\,{\fontsize{6}{7.2}\selectfont\textcolor{red!70!black}{$(-7.7)$}} & 50.0\,{\fontsize{6}{7.2}\selectfont\textcolor{red!70!black}{$(-4.7)$}} \\
\midrule
\multicolumn{7}{@{}l}{\textit{Retrieval (R@1)}} \\
Flickr30k T2I & 59.1 & 71.0 & 63.3 & 62.2 & 63.4\,{\fontsize{6}{7.2}\selectfont\textcolor{blue!70!black}{$(+4.3)$}} & 61.5\,{\fontsize{6}{7.2}\selectfont\textcolor{blue!70!black}{$(+2.4)$}} \\
Flickr30k I2T & 78.5 & 85.0 & 75.5 & 73.7 & 76.0\,{\fontsize{6}{7.2}\selectfont\textcolor{red!70!black}{$(-2.5)$}} & 73.1\,{\fontsize{6}{7.2}\selectfont\textcolor{red!70!black}{$(-5.4)$}} \\
COCO T2I & 30.2 & 45.0 & 38.1 & 33.9 & 38.3\,{\fontsize{6}{7.2}\selectfont\textcolor{blue!70!black}{$(+8.1)$}} & 32.7\,{\fontsize{6}{7.2}\selectfont\textcolor{blue!70!black}{$(+2.5)$}} \\
COCO I2T & 50.4 & 59.0 & 49.6 & 43.6 & 50.1\,{\fontsize{6}{7.2}\selectfont\textcolor{red!70!black}{$(-0.3)$}} & 43.1\,{\fontsize{6}{7.2}\selectfont\textcolor{red!70!black}{$(-7.3)$}} \\
\bottomrule
\end{tabular*}
\end{table}

\begin{figure*}[pos=tbp]
	\centering
	\includegraphics[width=\textwidth]{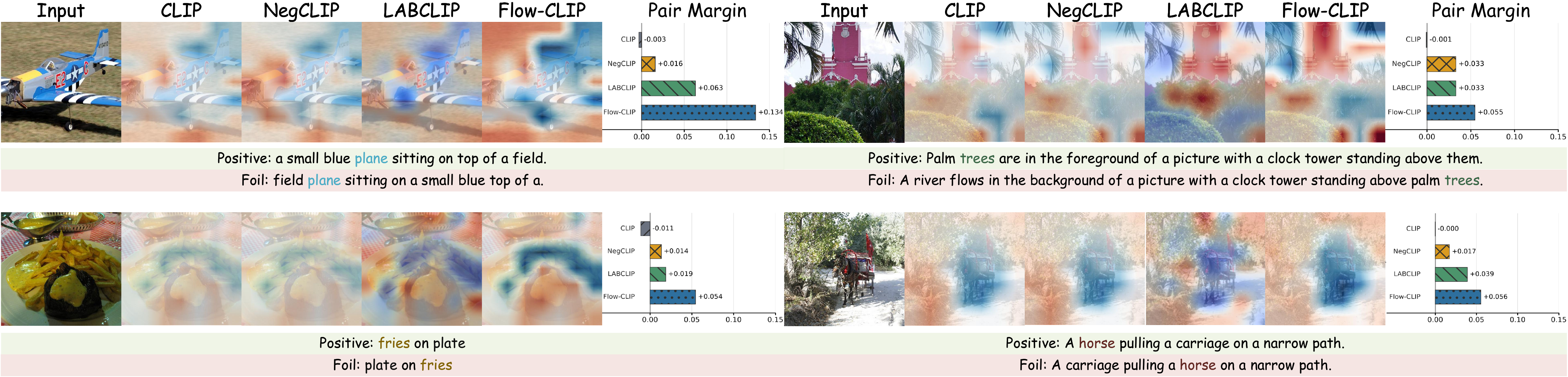}
	\caption{Compositional evidence across models. Word-order, object-replacement, relation-binding, and object-binding cases compare evidence maps and gold--foil score margins. Warm and cool colors indicate responses above and below the within-image reference. Margins are reported on each method's native scoring scale.}\label{fig:ranking-repairs}
	\vspace{6pt}
	\centering
	\includegraphics[width=0.98\textwidth]{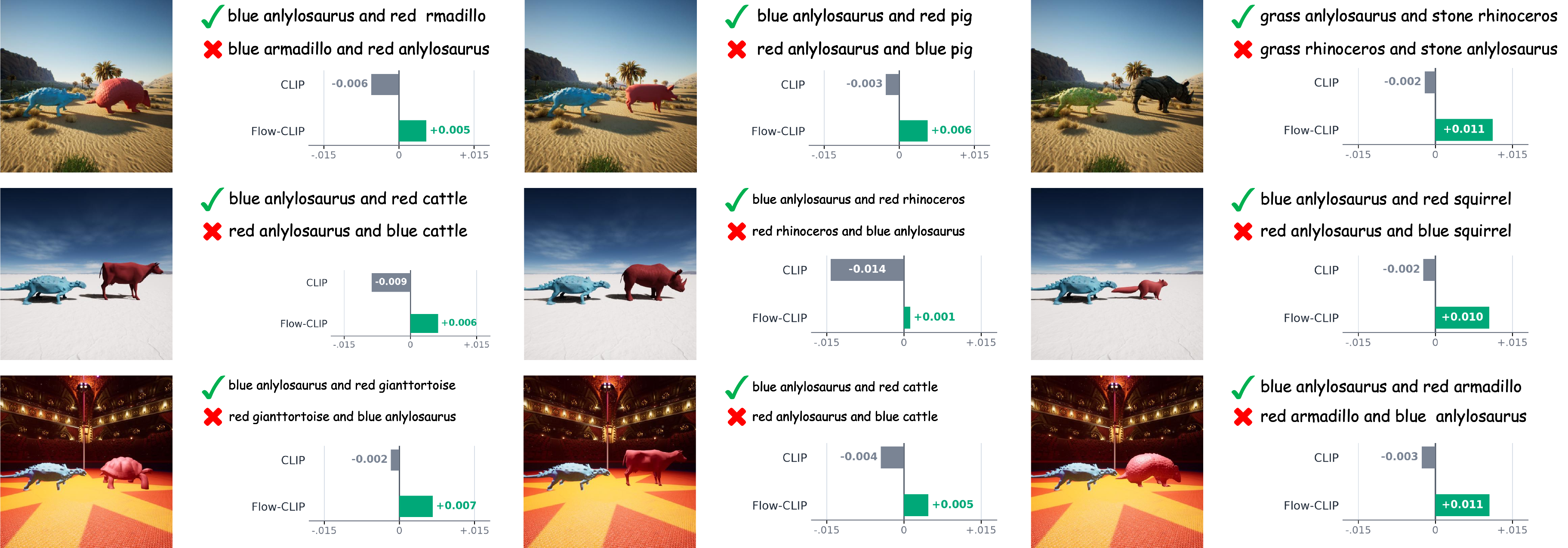}
	\caption{PUG:SPAR color--object binding repairs. Each example pairs an official
		PUG:SPAR image with a gold color--object assignment and a swapped foil. Bars show
		the gold--foil score margin for CLIP~\citep{radford2021learning} and Flow-CLIP. Positive values indicate that
		the gold binding is ranked higher. In these selected cases, Flow-CLIP ranks the gold assignment above the foil,
		reversing the CLIP decision.}\label{fig:pug-spar-binding}
	
\end{figure*}

The results in \Cref{tab:zero-shot-preservation} evaluate the learned score with class-name prompts and caption queries. Relative to pretrained CLIP~\citep{radford2021learning}, training on either COCO or CC3M improves text-to-image retrieval on Flickr30k and COCO, with larger gains after COCO training. Top-1 classification decreases on all three datasets, while COCO-trained Flow-CLIP nearly preserves COCO image-to-text retrieval.

Relative to Global-$M$, COCO-trained Flow-CLIP improves all four retrieval scores, whereas CC3M-trained Flow-CLIP reduces them. Classification changes vary across datasets. The retrieval effect thus differs between the COCO and CC3M adaptation settings, which also differ in caption generation, training budget and pooling temperature.

\begin{figure*}[pos=tp]
\centering
\fontsize{8}{9.6}\selectfont
\crepecomicfont
\setlength{\tabcolsep}{3pt}
\renewcommand{\arraystretch}{1.12}
\newcommand{\crepegoldcap}{\textcolor{green!45!black}{$\checkmark$}}
\newcommand{\crepefoilcap}{\textcolor{red!70!black}{$\times$}}
\begin{tabular}{@{}p{0.492\textwidth}@{\hspace{0.012\textwidth}}p{0.492\textwidth}@{}}
\toprule
\begin{tabular}[t]{@{}m{0.20\linewidth}>{\raggedright\arraybackslash}p{0.56\linewidth}cc@{}}
\multirow{7}{*}{\includegraphics[width=0.98\linewidth,height=0.105\textheight,keepaspectratio]{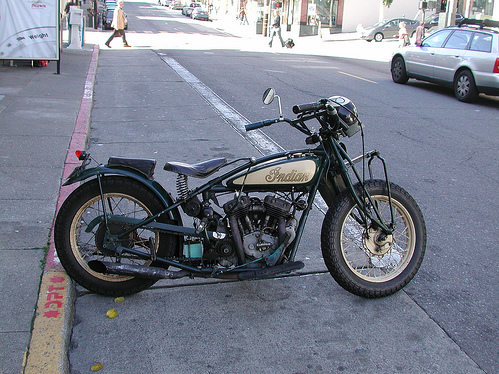}}
& Caption & CLIP & \shortstack{Flow-\\CLIP}\\
\cmidrule(l){2-4}
& \crepegoldcap{} cracks in pavement and another pavement & \#4 & \textbf{\#2}\\
& \crepefoilcap{} cracks across pavement and another pavement & \#3 & \#5\\
& \crepefoilcap{} mouths in pavement and another pavement & \#2 & \#6\\
& \crepefoilcap{} cracks in stand and another pavement & \#6 & \#3\\
& \crepefoilcap{} cracks under pavement and another pavement & \#5 & \#4\\
& \crepefoilcap{} dips in pavement and another pavement & \#1 & \#1\\
\end{tabular}
&
\begin{tabular}[t]{@{}m{0.20\linewidth}>{\raggedright\arraybackslash}p{0.56\linewidth}cc@{}}
\multirow{7}{*}{\includegraphics[width=0.98\linewidth,height=0.105\textheight,keepaspectratio]{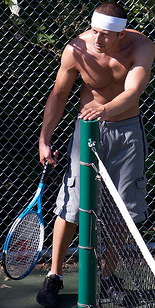}}
& Caption & CLIP & \shortstack{Flow-\\CLIP}\\
\cmidrule(l){2-4}
& \crepegoldcap{} man wearing gray shorts. there is a racket. & \#5 & \textbf{\#1}\\
& \crepefoilcap{} man wearing racket. there is a gray shorts. & \#1 & \#2\\
& \crepefoilcap{} shorts wearing gray man. there is a racket. & \#6 & \#6\\
& \crepefoilcap{} racket wearing gray shorts. there is a man. & \#3 & \#4\\
& \crepefoilcap{} gray shorts wearing man. there is a racket. & \#4 & \#5\\
& \crepefoilcap{} man wearing gray racket. there is a shorts. & \#2 & \#3\\
\end{tabular}
\\
\midrule
\begin{tabular}[t]{@{}m{0.20\linewidth}>{\raggedright\arraybackslash}p{0.56\linewidth}cc@{}}
\multirow{7}{*}{\includegraphics[width=0.98\linewidth,height=0.105\textheight,keepaspectratio]{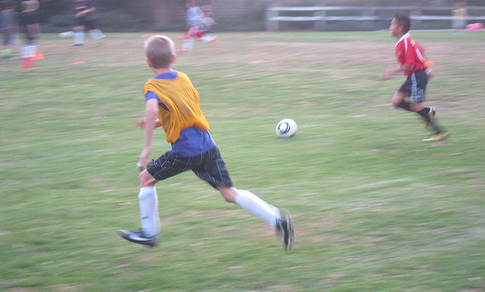}}
& Caption & CLIP & \shortstack{Flow-\\CLIP}\\
\cmidrule(l){2-4}
& \crepegoldcap{} boy playing with ball, ball2 in field & \#5 & \textbf{\#1}\\
& \crepefoilcap{} boy playing with ball, ball2 in right & \#3 & \#4\\
& \crepefoilcap{} boy playing with game, ball2 in field & \#4 & \#2\\
& \crepefoilcap{} boy playing with ball, ball2 in short & \#6 & \#3\\
& \crepefoilcap{} boy playbyg with ball, ball2 in field & \#2 & \#6\\
& \crepefoilcap{} player playing with ball, ball2 in field & \#1 & \#5\\
\end{tabular}
&
\begin{tabular}[t]{@{}m{0.20\linewidth}>{\raggedright\arraybackslash}p{0.56\linewidth}cc@{}}
\multirow{7}{*}{\includegraphics[width=0.98\linewidth,height=0.105\textheight,keepaspectratio]{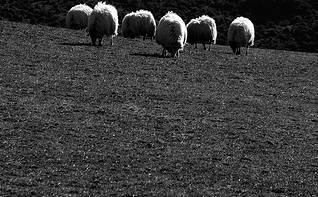}}
& Caption & CLIP & \shortstack{Flow-\\CLIP}\\
\cmidrule(l){2-4}
& \crepegoldcap{} sheep eating grass in a field & \#3 & \textbf{\#1}\\
& \crepefoilcap{} sheep eating peanut in a field & \#5 & \#5\\
& \crepefoilcap{} sheep eating bean in a field & \#4 & \#6\\
& \crepefoilcap{} sheep eatbyg grass in a field & \#2 & \#3\\
& \crepefoilcap{} sheep eatnearg grass in a field & \#1 & \#4\\
& \crepefoilcap{} sheep eating grass in a desert & \#6 & \#2\\
\end{tabular}
\\
\bottomrule
\end{tabular}
\caption{Caption rankings for four CREPE~\citep{ma2023crepe} examples. Each image has one gold caption and five foils. Columns give CLIP~\citep{radford2021learning} and Flow-CLIP ranks, with rank 1 best. Green checks identify gold captions and red crosses identify foils.}
\label{fig:crepe-six-candidate-ranking}
\end{figure*}

\subsection{Qualitative Analysis}\label{sec:visual-analysis}

\paragraph{\textbf{\upshape Local evidence.}}

The four examples in \Cref{fig:ranking-repairs} compare token--patch responses and gold--foil margins for word order, object replacement, relation binding and object binding. Warm and cool regions indicate responses above and below the within-image mean. They locate patch support for each caption, while the adjacent margins show which caption the final score prefers.

LABCLIP and Flow-CLIP both select the gold caption in all four pairs. Their respective margins are 0.063 and 0.134 for the aircraft example, and 0.019 and 0.054 for the plate example. Positive signs indicate correct pairwise rankings; the magnitudes are specific to each scoring function. The aircraft captions differ in word order, while the plate captions reverse an object relation.

\paragraph{\textbf{\upshape Attribute assignments.}}

The PUG:SPAR~\citep{bordes2023pug} examples illustrated in \Cref{fig:pug-spar-binding} hold the object inventory and scene templates fixed while exchanging color assignments. CLIP~\citep{radford2021learning} assigns a negative gold--foil score margin to every displayed pair, whereas Flow-CLIP assigns a positive margin.

\begin{figure*}[pos=tbp]
\centering
\includegraphics[width=\textwidth]{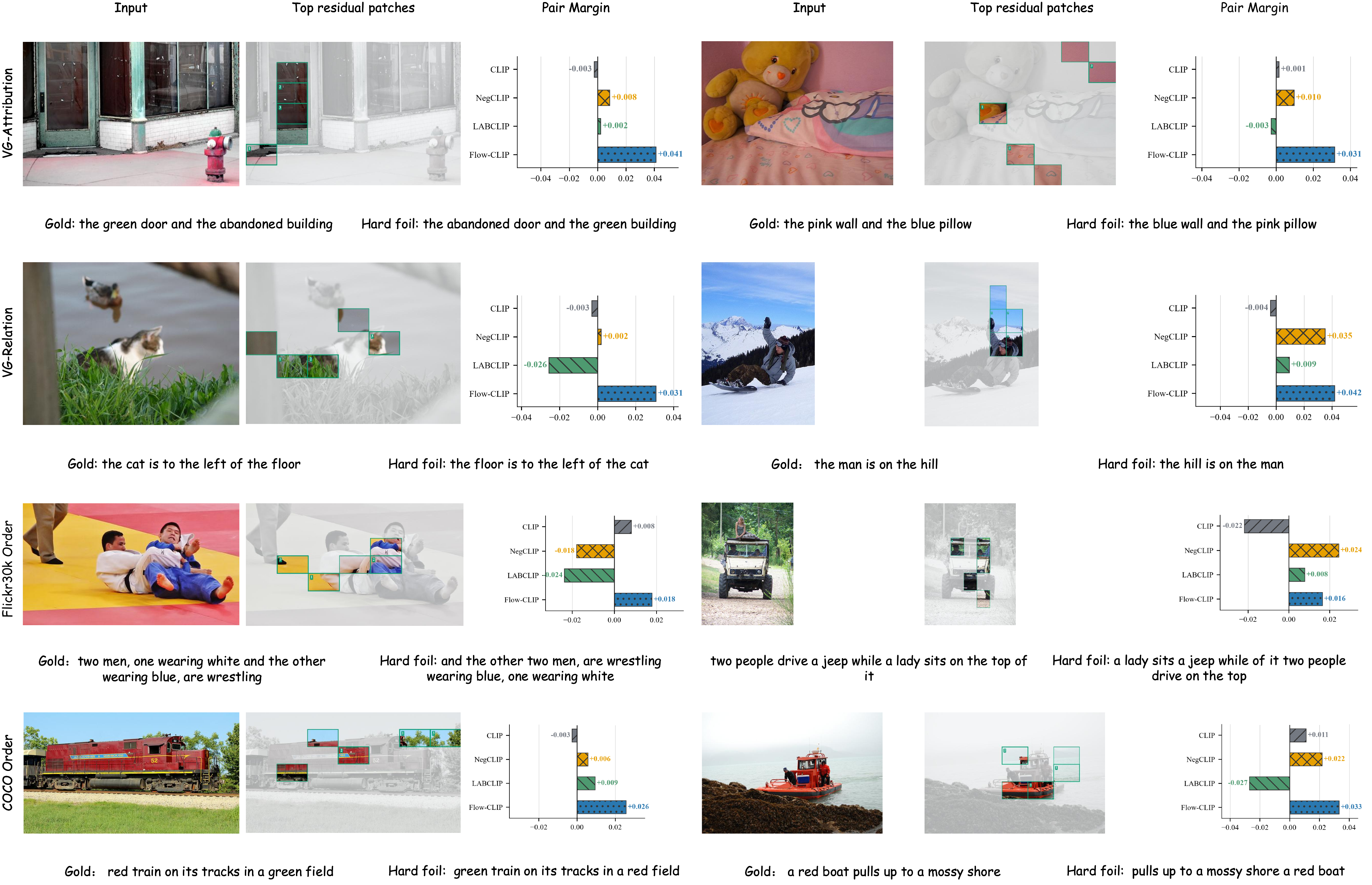}
\caption{Cross-model residual-evidence atlas on ARO. The eight examples cover
VG-Attribution, VG-Relation, Flickr30k Order, and COCO Order. For each example,
the input image is followed by patches selected by Flow-CLIP's centered-feature sensitivity
and the gold--hard-foil score margins of CLIP~\citep{radford2021learning}, NegCLIP~\citep{yuksekgonul2023and}, LABCLIP~\citep{koishigarina2025clip}, and Flow-CLIP.
Positive margins indicate correct pairwise rankings.}\label{fig:cross-model-residual-atlas}
\end{figure*}

\begin{samepage}
\paragraph{\textbf{\upshape Caption rankings.}}

In the four CREPE~\citep{ma2023crepe} examples in \Cref{fig:crepe-six-candidate-ranking}, Flow-CLIP selects the gold caption in three cases where CLIP~\citep{radford2021learning} selects a foil.
\par\end{samepage}

In the racket scene, the selected caption correctly identifies the man's clothing, while CLIP prefers a description that treats the racket as clothing. In the football and sheep scenes, Flow-CLIP ranks the gold descriptions above alternatives that substitute the described object or location.

The pavement example shows a partial improvement: the gold caption rises from fourth to second, but an alternative describing a different pavement feature remains first. The higher gold-caption rank still leaves another candidate as the final selection.

\paragraph{\textbf{\upshape Patch sensitivity.}}

The eight cases in \Cref{fig:cross-model-residual-atlas} compare patch sensitivity and caption-pair margins across attribution, relation and word-order tasks. Patch selection uses the first-order change in the gold--foil evidence increment along the centered patch-feature direction, highlighting regions to which this local comparison is sensitive.

\begin{samepage}
In the attribution cases, the selected regions cover the door, building, wall and pillow whose properties distinguish the captions. The adjacent margins report the final caption preference.
\par\end{samepage}

\paragraph{\textbf{\upshape Margin changes.}}

\begin{figure}[pos=htbp]
	\centering
	\includegraphics[width=\columnwidth]{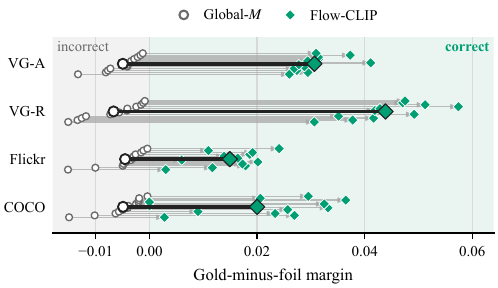}
	\caption{Changes in caption-pair margins. Each arrow connects the Global-$M$ and Flow-CLIP gold--foil score margins for the same image and caption pair. The zero boundary separates incorrect and correct rankings. Enlarged markers show the task means.}
	\label{fig:residual-decision-flow}
\end{figure}

In \Cref{fig:residual-decision-flow}, gold--foil margins increase for 47 of 48 illustrative pairs and decrease for one. After excluding two order cases with duplicate gold captions, correct top-1 selections increase from 17 to 45 of 46 cases, with 28 repairs and no correct-to-incorrect reversals. A COCO Order case remains incorrect: its gold rank worsens from 3 to 4 as another caption outranks it despite the improved margin over the displayed foil. These order-task candidate sets differ from the full benchmark.

\AlignedFloatBarrier
\section{Conclusion}\label{sec:conclusion}

We presented Flow-CLIP to improve attribute--object binding and mitigate reliance on language priors in compositional image--text matching. GCC calibrates global alignment, ICER adjusts this score through image-conditioned changes in token--patch evidence, and CIMR supervises caption-pair margins across matched and replacement images. Experiments show improved binding beyond learned global alignment and reduced reliance on language priors. These findings support using image-conditioned evidence refinement to resolve binding ambiguities that persist after sentence-level alignment.

\section*{AI USE STATEMENT}

Generative AI tools were used only to assist with manuscript writing.

\bibliographystyle{elsarticle-num}
\setlength{\parskip}{0pt}
\renewcommand{\bibfont}{\fontsize{8pt}{9.4pt}\selectfont}
\setlength{\bibsep}{0pt plus 0.25pt}

\bibliography{references}
\end{document}